\documentclass[11pt]{article}

\usepackage[final]{acl}

\usepackage{times}
\usepackage{latexsym}
\usepackage{amsmath}
\usepackage{amsfonts}
\usepackage{amssymb}
\usepackage{sparklines}
\usepackage[T1]{fontenc}

\usepackage[utf8]{inputenc}

\usepackage{microtype}

\usepackage{inconsolata}

\usepackage{graphicx}
\usepackage{booktabs}   
\usepackage{algorithm}
\usepackage{colortbl}
\usepackage{xcolor}
\usepackage{tcolorbox}
\usepackage{multirow}
\usepackage{tikz}
\usepackage{algpseudocode}
\title{BEiTScore++: Reference-free Image Captioning Evaluation with BEiT-3 and Conformalized Quantile Regression}

\author{
  \textbf{Gonçalo Gomes\textsuperscript{1,2}},
  \textbf{Bruno Martins\textsuperscript{1,2}},
  \textbf{Chrysoula Zerva\textsuperscript{3,4}},
\\
  \textsuperscript{1}Instituto Superior Técnico, University of Lisbon
\\
  \textsuperscript{2}INESC-ID; 
\quad
  \textsuperscript{3}Instituto de Telecomunicações
\\
\textsuperscript{4}AILS Lab; ECE, National Technical University of Athens
\\
  \small{
    {\{goncaloecgomes, chrysoula.zerva, bruno.g.martins\}@tecnico.ulisboa.pt}, czerva@mail.ntua.gr
  }
}

\begin{document}
\maketitle
\begin{abstract}
Reliable uncertainty estimation is important for deploying evaluation metrics in real-world settings, where both input ambiguity and domain shifts are common. In this work, we propose an uncertainty-aware extension of the BEiTScore image captioning metric, producing conformally calibrated confidence intervals alongside alignment scores. Our approach models predictive distributions over human judgments, enabling both accurate scoring and principled uncertainty quantification in a single forward pass. We evaluate our BEiTScore++ metric across multiple benchmarks with human-annotated ratings, showing that it preserves the ranking correlation of BEiTScore, while producing intervals whose width tracks the prediction error more closely than Monte Carlo dropout on most benchmarks. The estimated uncertainty also grows under image degradation and on out-of-domain artwork images, making BEiTScore++ a practical tool for uncertainty-aware evaluation of image-caption alignment.
\end{abstract}

\vspace{-0.13cm}
\section{Introduction}
\vspace{-0.13cm}

Image captioning evaluation models typically produce a single score that captures the semantic alignment between the image content and a textual description, aiming at scores that are highly correlated with human assessments. However, individual point estimates make it challenging to convey how confident the model is in the produced assessments. This limitation becomes particularly relevant in scenarios where multiple captions receive similar scores while differing substantially in reliability. In these cases, an effective evaluation metric should not only predict caption quality, but also provide an informative estimate of uncertainty.

\begin{figure}[t]
    \centering
    \includegraphics[trim={0 0cm 0 0},clip,width=\linewidth]{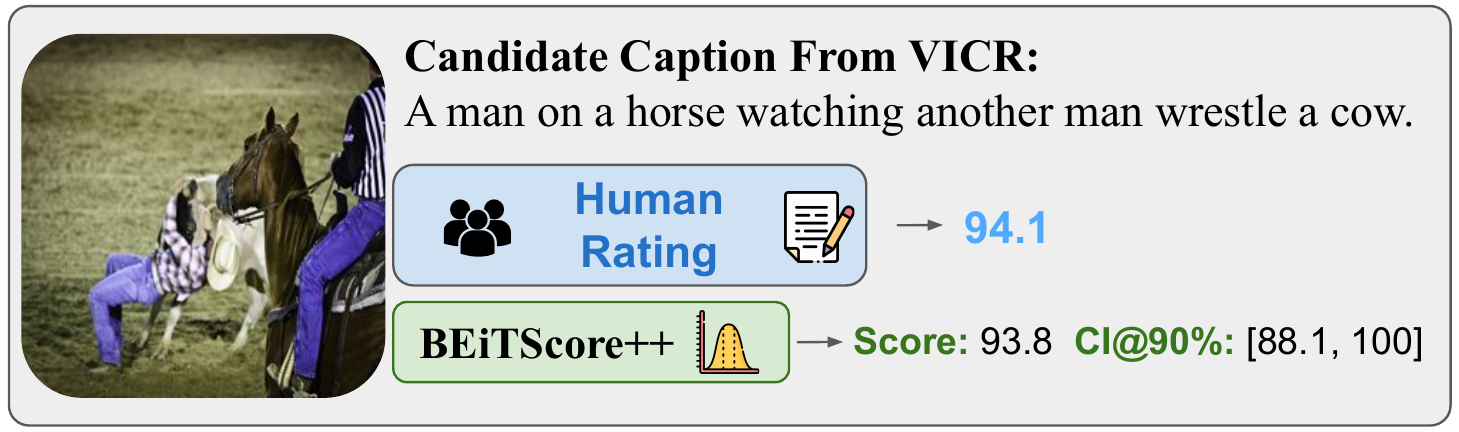}
    \vspace{-0.13cm}
    \caption{Qualitative example for the use of BEiTScore++ in an instance from the VICR~\cite{narins2023validated} benchmark. The candidate caption is scored as highly relevant by humans, as well as by BEiTScore++. Moreover, BEiTScore++ also outputs a confidence interval, covering with high probability the score assigned by humans.}
    \vspace{-0.13cm}
    \label{fig:intro_fig}
\end{figure}

This work introduces BEiTScore++ as an uncertainty-aware extension to the state-of-the-art BEiTScore image captioning metric~\cite{beitscore}, which predicts both a quality score and a calibrated confidence interval. We formulate the evaluation problem as quantile regression and train the model using a centered interval objective that encourages both coverage and sharpness. To ensure coherent intervals, the quantiles are constructed monotonically from a base quantile prediction, avoiding quantile crossing by design. Moreover, to improve coverage reliability, we further apply post-hoc conformal calibration. Figure~\ref{fig:intro_fig} provides a general illustration of the proposed procedure.\looseness-1

BEiTScore++ is evaluated on several benchmark datasets with human numeric ratings, allowing us to assess how well the predicted uncertainty correlates with prediction error and human judgments. We show that correlation with human judgments remains at the level of state-of-the-art evaluation metrics, and that the width of the intervals correlates with prediction error. After calibration at a 90\% target, coverage ranges from 81\% to 98\% on in-domain benchmarks, but remains lower on out-of-distribution data.

Beyond predictive performance, we further investigate whether the estimated uncertainty reflects meaningful sources of variability. We focus on two key factors: input ambiguity and model knowledge gaps. To study input ambiguity, we introduce controlled image degradations, such as Gaussian blur, and observe that uncertainty increases under moderate degradation. To examine model knowledge gaps, we evaluate the model on out-of-domain artwork datasets and find that uncertainty rises compared to in-domain natural images, consistent with epistemic uncertainty due to domain shift, although artwork is also intrinsically more ambiguous.
Overall, BEiTScore++ aims to make image captioning evaluation not only more accurate, but also more trustworthy.\footnote{The corresponding source code and the different models are available at \url{https://github.com/gecgomes/Uncertainty_BEiTScore_pp}}

\vspace{-0.13cm}
\section{Related Work}
\vspace{-0.13cm}

This section reviews prior work concerning three key areas: image captioning evaluation, uncertainty quantification, and uncertainty calibration.

\vspace{-0.13cm}
\subsection{Image Captioning Evaluation}
\vspace{-0.13cm}

CLIPScore~\cite{hessel2021clipscore} was proposed to overcome the limitations of reference-dependent evaluation methods by enabling reference-free scoring through the cosine similarity between CLIP image and text embeddings. However, the CLIP text encoder typically imposes a strict token limit (e.g., 77 tokens), which restricts its use for long-form descriptions. Several extensions and related approaches have been proposed~\cite{sarto2025positive,hu2023infometic,narins2023validated,wada2024polos,gomes2025evaluation}, including approaches that adapt CLIP-based methods to longer textual inputs~\cite{chen2025specs,zhang2024long}.

Other approaches leverage Large Language Models (LLMs) for caption evaluation~\cite{chan2023clair, lee2024fleur, kim2025expert}. These models benefit from strong zero-shot capabilities, long context windows, and the ability to capture nuanced semantic differences. However, they incur high computational costs and slow inference times, making them impractical for large-scale benchmarking or integration into decoding pipelines.

More recent work revisits learned encoder-based metrics using more expressive cross-encoder architectures. ~\citet{beitscore} proposed BEiTScore, with basis on a cross-encoder initialized from a BEiT-3 visual question answering checkpoint, this way combining strong pretrained representations with computational efficiency. BEiTScore employs cross-attention between the visual and textual modalities, enabling fine-grained sensitivity to visual-linguistic inconsistencies. Its training strategy incorporates a diverse data mixture designed to expose the model to long descriptions and complex visual-linguistic relationships, along with adversarial augmentations that enhance compositional coverage and relational reasoning. BEiTScore matches or surpasses specialized LLM-based metrics on most benchmarks, at a fraction of their computational cost.

\vspace{-0.13cm}
\subsection{Uncertainty Quantification}
\vspace{-0.13cm}

Applications that rely on machine learning predictions often require an estimate of confidence alongside the prediction itself, in order to mitigate overconfident decisions. The process of estimating this confidence is commonly referred to as uncertainty quantification. From a philosophical perspective, uncertainty can be framed by asking \emph{where does the randomness lie?} This leads to two distinct viewpoints, namely Bayesian approaches which attribute randomness to model parameters, and frequentist approaches which assume randomness originates from the data.

In Bayesian methods, neural network weights are modeled as probability distributions rather than fixed values~\cite{mackay1992bayesian,welling2011bayesian,tran2019bayesian}. Training focuses on learning these distributions and, during inference, weights are sampled accordingly. While theoretically appealing, these approaches are computationally expensive, particularly for modern models with millions or billions of parameters, leading to the widespread adoption of cheaper approximations.

A common approximation-based approach corresponds to deep ensembles, where multiple models are trained independently and the empirical variance of their predictions is used as a measure of uncertainty~\cite{lakshminarayanan2017simple,pearce2020uncertainty,zhan2023test}. Another widely used technique is Monte Carlo dropout~\cite{gal2016dropout,glushkova2021uncertainty}, an approximation to Bayesian inference that applies dropout at test time to perform multiple stochastic forward passes, estimating uncertainty through variations in the outputs. However, both approaches require multiple forward passes at inference time to construct uncertainty estimates, which can become increasingly slow as the number of samples grows.

Beyond variance-based approaches, quantile regression has emerged as a compelling alternative, particularly in settings where the model can be fully trained and controlled. Quantiles offer a flexible and non-parametric representation of uncertainty, are directly interpretable in the target space, and enable straightforward construction of prediction intervals. Traditionally, quantile regression is trained using the pinball loss~\cite{koenker2001quantile}. However, recent work~\cite{chung2021beyond} shows that the trade-off between sharpness and coverage that is induced by the pinball loss is implicit and arbitrary, observing empirically that, in highly expressive models, it tends to favor sharpness at the expense of calibration. To address this limitation, \citet{chung2021beyond} propose alternative loss functions that explicitly balance sharpness and coverage calibration during training. 

\vspace{-0.13cm}
\subsection{Uncertainty Calibration}
\vspace{-0.13cm}

Uncertainty calibration naturally complements uncertainty quantification, as confidence estimates are only meaningful if they reflect the true frequency with which events occur. In the context of prediction intervals, calibration requires that nominal confidence levels match empirical coverage, for example aiming toward a \(90\%\) interval that contains the true value approximately \(90\%\) of the time. 

For quantile-based methods, a key prerequisite for well-calibrated intervals is to avoid \emph{quantile crossing}, where lower quantiles exceed higher ones and the induced prediction sets become incoherent. ~\citet{narayan2021regularization} address this by enforcing monotonicity in the quantile level within deep lattice networks, guaranteeing non-crossing quantiles while retaining model flexibility.

Conformal Risk Control (CRC) provides a general, distribution-free framework for calibrating arbitrary user-defined risk functions. ~\citet{angelopoulos2022conformal} extend split conformal prediction by choosing a scalar parameter that controls the expected value of any bounded loss that is monotone in the size of the prediction set. When the loss is chosen as an indicator of miscoverage, CRC recovers classical conformal prediction with finite-sample coverage guarantees, while more general loss choices enable direct control of application-specific risks, such as the false negative rate or task-specific distortion measures.

Recent work has, for instance, leveraged this framework to improve both granularity and calibration of image captioning evaluation metrics. ~\citet{gomes2025conformal} generate distributions over CLIPScore values via simple perturbation strategies (e.g., input masking), and then apply conformal risk control to calibrate these score distributions for word-level error detection and to construct calibrated CLIPScore intervals. This approach yields formal guarantees on user-specified risk bounds, detects erroneous words in captions, and improves correlation between uncertainty estimates and residuals, thus enhancing CLIPScore's reliability as an image captioning evaluation metric.

\vspace{-0.13cm}
\section{Properties of Uncertainty Estimates}
\vspace{-0.13cm}

Uncertainty estimates should satisfy several desirable properties that make them both statistically valid and practically useful. In this work, we focus on four central criteria: calibration, sharpness, monotonicity, and conditional coverage. We also emphasize interpretability and robustness to distribution shift, since uncertainty is only useful if it can be understood and trusted in deployment.

Calibration means that nominal confidence levels match empirical frequencies: for example, a \(90\%\) prediction interval should contain the true target about \(90\%\) of the time. In turn, sharpness refers to the informativeness of the estimate, typically measured by the width of the prediction interval or the dispersion of the predictive distribution. Following standard principles in probabilistic forecasting, one should first aim to achieve valid coverage and then, among calibrated estimators, maximize sharpness~\cite{gneiting2007strictly, chung2021beyond}. In other words, narrow intervals are desirable only when they remain reliable.

For quantile-based uncertainty estimates, the predicted quantiles should be monotone in the quantile level. This prevents quantile crossing, where a lower quantile exceeds a higher one, rendering the resulting interval  incoherent. Enforcing non-crossing is important not only for mathematical consistency, but also for producing interpretable confidence intervals that can be reliably used downstream~\cite{narayan2021regularization}.

A calibrated uncertainty estimate may satisfy marginal coverage, meaning that the desired coverage holds on average over the full data distribution. However, marginal validity alone can hide systematic failures on particular regions of the input space or on minority subgroups~\cite{angelopoulos2021gentle, zervaconformalizing}. Conditional coverage is therefore a stronger and more desirable property, requiring reliable coverage within subpopulations or for individual types of inputs. However, conditional coverage is something that, in most cases, is very hard if not impossible to achieve~\cite{vovk2012conditional}, and thus a relaxation of this property is often targeted.

Uncertainty estimates should also be easy to interpret, communicating not only how uncertain the model is, but also what the uncertainty means in the scale of the task, which supports human review and downstream risk-aware decisions.

Finally, a useful uncertainty estimate should be sensitive to different types of inputs. In particular, uncertainty should typically increase when the model encounters inputs that are highly ambiguous, or that differ from the training distribution, rather than having outputs that remain overconfident.

\vspace{-0.13cm}
\section{An Uncertainty Aware BEiTScore} \label{sec:beitscore++}
\vspace{-0.13cm}
\label{sec:method}

\begin{figure*}
    \centering
    \includegraphics[width=0.85\linewidth]{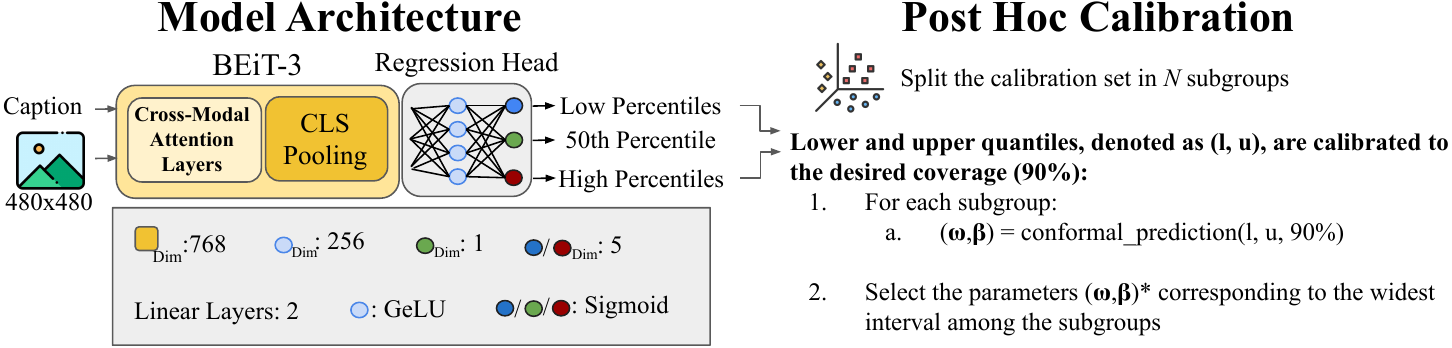}
\vspace{-0.15cm}
    \caption{The proposed architecture for uncertainty modeling and the respective post-hoc calibration method.}
    \vspace{-0.15cm}
    \label{fig:uncertainty_beitscore}
\end{figure*}

\begin{algorithm}[t]
\caption{Monotone quantile construction for uncertainty-aware predictions in BEiTScore++}
\label{alg:monotone_quantiles}
\begin{algorithmic}[1]
\Require Image $I$ and text description $T$
\State $\mathbf{h} \gets \mathrm{BEiT3}(I, T)$
\State Predict median and bounded deltas:
\State \hspace{0.5cm} $q_{\mathrm{mid}} \gets \sigma(f_{\mathrm{mid}}(\mathbf{h}))$
\State \hspace{0.5cm} $\boldsymbol{\delta}_{\mathrm{low}} \gets \sigma(f_{\mathrm{low}}(\mathbf{h})) \in (0,1)^{K_{\mathrm{low}}}$
\State \hspace{0.5cm} $\boldsymbol{\delta}_{\mathrm{high}} \gets \sigma(f_{\mathrm{high}}(\mathbf{h})) \in (0,1)^{K_{\mathrm{high}}}$
\State Initialize quantile list: $\mathbf{q} \gets [\,q_{\mathrm{mid}}\,]$
\Statex \textit{// Build lower tail quantiles by prepending} 

\State \textbf{for }{$i = 1$ to $K_{\mathrm{low}}$}\textbf{do } \\
    $\quad q_{\mathrm{new}} = \delta_{\mathrm{low}}^{(K_{\mathrm{low}}-i+1)} \cdot \mathbf{q}[0]$ \\
    \hspace{0.3cm} Prepend $q_{\mathrm{new}}$ to $\mathbf{q}$

\Statex \textit{// Build upper tail quantiles by appending}

\State \textbf{for }{$j = 1$ to $K_{\mathrm{high}}$}\textbf{do } \\
    $\quad q_{\mathrm{new}} = \mathbf{q}[-1] + \delta_{\mathrm{high}}^{(j)} \cdot (1 - \mathbf{q}[-1])$ \\
    \hspace{0.3cm} Append $q_{\mathrm{new}}$ to $\mathbf{q}$

\State \Return $\mathbf{y} \gets \mathbf{q}$
\end{algorithmic}
\end{algorithm}

We formulate uncertainty estimation for BEiTScore as a quantile regression problem. The proposed model, which we named BEiTScore++, combines a BEiT-3 backbone with a quantile regression head comprising 11 output nodes, which predict five centered prediction intervals (70\%, 75\%, 80\%, 85\%, 90\%) together with a central value corresponding to the 50th percentile. These intervals provide a nested family of ``typical'' uncertainty bands around the median. Jointly learning this grid of quantiles encourages the model to capture a smooth, coherent approximation of the full conditional distribution of scores, which acts as a form of distributional regularization compared to training independent models for individual quantiles~\cite{narayan2021regularization}. 

\paragraph{Quantile Regression Loss:} Rather than relying on the standard pinball loss, we use the Centered Interval Loss proposed by \citet{chung2021beyond}, which explicitly balances coverage calibration and interval sharpness. We denote a $(1-\alpha)$ confidence interval by $\hat{l}_{\alpha} = \mathbb{Q}_{\frac{\alpha}{2}}$ and $\hat{u}_{\alpha} = \mathbb{Q}_{1-\frac{\alpha}{2}}$, and define the centered interval loss as:
\begin{equation}
\resizebox{0.89\columnwidth}{!}{$
    \mathcal{L}^\alpha_{\text{CI}} = (\hat{u}_\alpha - \hat{l}_\alpha) + \frac{2}{\alpha}(\hat{l}_\alpha - y)\,\mathbb{I}\{y < \hat{l}_\alpha\} + \frac{2}{\alpha}(y - \hat{u}_\alpha)\,\mathbb{I}\{y > \hat{u}_\alpha\}.
$}
\end{equation}
This loss explicitly penalizes the width of the intervals, thus promoting sharpness. \citet{chung2021beyond} show that the minimum of its expected value is attained at the true conditional quantiles, thereby also encouraging well-calibrated coverage. The considered design reflects the goal of probabilistic forecasting: intervals should be not only narrow but also correctly calibrated. Accordingly, the lower and upper quantiles are trained jointly to improve interval quality, while the median estimate is supervised with an $L_1$ loss. The full training objective is thus the sum of the centered interval losses across the five intervals, plus the $L_1$ loss on the median, defined as follows: 
\begin{equation}
\mathcal{L}_{\text{total}} 
= \sum_{\alpha \in \mathcal{A}} \mathcal{L}^\alpha_{\text{CI}}
  +  \bigl| y - \mathbb{Q}_{0.5} \bigr|,
\end{equation}
where $\mathcal{A} = \{0.10, 0.15, 0.20, 0.25, 0.30\}$ corresponds to the $(90\%, 85\%, 80\%, 75\%, 70\%)$ prediction intervals. Figure~\ref{fig:uncertainty_beitscore} illustrates the proposed architecture for uncertainty modeling, and the corresponding post-hoc calibration procedure.

\paragraph{Monotone Quantile Construction.} The middle quantile in BEiTScore++, i.e., the 50th percentile, is predicted directly by the respective node, and the remaining quantiles are constructed iteratively from the center by predicting positive deltas towards the lower and upper tail. Concretely, the model first estimates the central score, and then progressively expands toward lower and higher quantiles using bounded multiplicative offsets. This design enforces quantile monotonicity by construction, thus preventing quantile crossing and ensuring that the predicted intervals remain ordered. In addition, because all quantile outputs are constrained to lie in \([0,1]\), they remain compatible with the score range of BEiTScore and preserve interpretability. Algorithm~\ref{alg:monotone_quantiles} summarizes the monotone quantile construction used in BEiTScore++.

\paragraph{Quality Score Estimation.} The regression head that predicts both the median and the other quantile estimates are initialized from the original BEiTScore model, which was trained with a diverse data mixture designed to expose the model to long descriptions and complex visual-linguistic relations. For the nodes that estimate the lower and upper deltas, we additionally subtract 2.5 from the bias inherited from the original BEiTScore regression head. Since every delta is a sigmoid output, all deltas start small, so the lower quantiles start near 0 while the upper quantiles start close to the median. This deliberately asymmetric initialization was chosen empirically because it yields a plausible coverage before training, from which training can progressively refine calibration and sharpness on both sides of the interval. This initialization follows the principle that probabilistic forecasts should first achieve calibration and then optimize sharpness, as emphasized by \citet{gneiting2007strictly}. Empirically, we found that initializing all quantile heads identically and aiming at the generation of small values (i.e., maximum sharpness), without this specific change in the bias, led to a suboptimal local minimum, producing intervals that were very sharp but poorly calibrated, with weaker correlation between prediction error and estimated uncertainty.

\paragraph{Conformalizing Intervals.} We apply a post-hoc calibration step using conformal risk control. We define the calibrated prediction interval as follows:
\begin{align}
    \hat{l} &= \max\bigl(0, (l-m)\cdot\omega + m - \beta\bigr), \label{eq:cal_lower} \\
    \hat{u} &= \min\bigl(1, (u-m)\cdot\omega + m + \beta\bigr), \label{eq:cal_upper}
\end{align}
where \(l\), \(m\), and \(u\) denote the lower, median, and upper quantile predictions that are being targeted for conformalization, respectively, and where \(\omega\) and \(\beta\) are calibration parameters optimized within the conformal framework to match a desired coverage level. Our procedure follows a general conformal risk control formulation~\cite{angelopoulos2022conformal,gomes2025conformal}, where the calibration objective is to satisfy a user-specified risk constraint, in our case corresponding to miscoverage.

\paragraph{Approximating Conditional Coverage:} As noted by \citet{angelopoulos2021gentle}, conformal calibration on a held-out calibration set provides formal guarantees only for marginal coverage. While this is valuable, it may hide poor behavior on specific subsets of the data. To favor coverage across different regions of the input space, we group training and calibration examples according to image similarity. Specifically, we extracted image embeddings and apply \(k\)-means clustering to partition the data into homogeneous subgroups. The number of clusters (i.e., $k$ = 63 in our tests) was selected using a combination of the elbow method and the silhouette score. We then partitioned the training and calibration set according to these clusters.

With the training dataset partitioned into subgroups, we organize mini-batches so that, at each iteration, all samples in a batch are drawn from a single subgroup selected at random. This design is inspired by the group-batching strategy used to induce adversarial group calibration in \citet{chung2021beyond}, where datapoints are grouped based on input features and gradient updates are computed on group-specific batches to improve adversarial group calibration. In our setting, we additionally use the partitioned calibration set to perform conformal calibration independently within each subgroup, obtaining $k$ candidate pairs of calibration parameters $(\omega, \beta)$. We then select the most conservative pair (i.e., the one producing the largest mean calibrated width over the calibration set), so that the coverage target is met within each calibration cluster. This rule prioritizes coverage over sharpness, and Appendix~\ref{app:cluster-sweep} shows that its effect comes mostly from the wider intervals it selects.

\vspace{-0.13cm}
\section{Experimental Evaluation}
\vspace{-0.13cm}

This section starts by presenting the evaluation methodology, describing implementation details, datasets, and evaluation metrics. It then describes the experimental results, starting with tests that assessed the correlation to human ratings, and then analyzing the uncertainty estimates. 

\vspace{-0.13cm}
\subsection{Implementation Details}
\vspace{-0.13cm}

We implement BEiTScore++ using a simple regression head on top of BEiT-3 encoders of size base and large. The regression head has a single hidden layer with 256 units and GeLU activations, as in BEiTScore. The encoder weights, together with the output nodes, are initialized from the original BEiTScore model. In turn, 2.5 is additionally subtracted from the bias of the lower and upper delta nodes, as described in Section~\ref{sec:beitscore++}. This value was chosen empirically so that the quantiles start with a coverage closer to their respective targets. The complete output layer consists of 11 nodes with sigmoid activations. The model is trained with a batch size of 16 using early stopping with a patience of 15 epochs. All experiments were conducted on a single NVIDIA H200 GPU, although comparable results can be obtained on lesser-memory GPUs by reducing the batch size and using gradient accumulation accordingly.

\vspace{-0.13cm}
\subsection{Datasets and Evaluation Metrics}
\vspace{-0.13cm}

Model training uses a mixture of datasets with human-annotated image-caption alignment scores, including VICR~\cite{narins2023validated}, Polaris~\cite{wada2024polos}, Nebula~\cite{matsuda2024deneb}, and EvalMuse-40k~\cite{han2024evalmuse}. These datasets provide continuous human judgments which we normalize to the range [0,1].

To evaluate predictive performance and the quality of uncertainty estimates, we conduct experiments on several benchmark datasets with human ratings. These include VICR~\cite{narins2023validated}, Polaris~\cite{wada2024polos}, Nebula~\cite{matsuda2024deneb}, Flickr8K-Expert~\cite{hodosh2013framing}, and Composite~\cite{aditya2015images}. In addition, for the analysis of different sources of uncertainty, we consider additional datasets spanning in-domain and out-of-domain distributions. In-domain datasets include IIW~\cite{garg2024imageinwords}, DOCCI~\cite{onoe2024docci}, Localized Narratives~\cite{pont2020connecting}, and MSCOCO~\cite{lin2014microsoft}. In turn, out-of-domain datasets focus on artwork images, namely with Docent~\cite{ananthram2025posh} and ArtCaps~\cite{lu2022artcap}. Appendix~\ref{app:datasets} provides a detailed characterization of all the datasets that were used in the experiments.

We assess model performance using four complementary metrics: coverage,
Kendall's $\tau_c$, sharpness, and UPS. Coverage evaluates whether predicted
intervals achieve the desired confidence level. Kendall's $\tau_c$ measures
ordinal agreement between median predictions and human ratings, being widely
used in prior work focused on image captioning evaluation metrics. Sharpness
quantifies the tightness of the predicted intervals, with narrower intervals
indicating more informative uncertainty estimates. Given the predicted lower
and upper interval bounds $l_i$ and $u_i$ for each of the $N$ evaluated
samples, sharpness is computed as the mean interval width,
\begin{equation}
    \mathrm{Sharpness} = \frac{1}{N} \sum_{i=1}^{N} \left( u_i - l_i \right).
\end{equation}
Notice that this measure is only meaningful when the predicted quantiles are monotone. In
the presence of quantile crossing, where $l_i > u_i$ for some samples, the
width $u_i - l_i$ becomes negative for these samples and no longer corresponds to a valid
interval. For these samples sharpness ceases to be a reliable measure of informativeness and can be driven artificially low by crossed predictions. 

Finally,
the Uncertainty Pearson Score (UPS) measures the correlation between predicted
uncertainty and prediction error, capturing whether higher uncertainty aligns
with larger deviations from human judgments. Taking the interval width
$\sigma_i = u_i - l_i$ as the uncertainty estimate and
$e_i = \lvert m_i - y_i \rvert$ as the absolute prediction error, where
$m_i$ is the 50th quantile and $y_i$ is the human rating,
UPS is defined as the Pearson correlation between the two:
\begin{equation}
    \mathrm{UPS} = \rho\left( e, \sigma \right),
\end{equation}
where $\rho(\cdot, \cdot)$ denotes the Pearson correlation coefficient, and
$e = (e_1, \dots, e_N)$ and $\sigma = (\sigma_1, \dots, \sigma_N)$ collect the
per-sample errors and uncertainties. Higher UPS values indicate that the model
assigns greater uncertainty precisely to the samples on which its predictions
are least accurate.

\vspace{-0.13cm}
\subsection{Assessments Using Human Ratings}
\vspace{-0.13cm}

\begin{table*}[t]
\centering
\resizebox{\textwidth}{!}{
\begin{tabular}{llll
cccc
cccc
cccc
cccc
cccc}
\toprule
\multicolumn{4}{l}{Prior to Calibration}
& \multicolumn{4}{c}{VICR}
& \multicolumn{4}{c}{Polaris}
& \multicolumn{4}{c}{Nebula}
& \multicolumn{4}{c}{Flickr8K-Expert}
& \multicolumn{4}{c}{Composite} \\
\cmidrule(lr){5-8}
\cmidrule(lr){9-12}
\cmidrule(lr){13-16}
\cmidrule(lr){17-20}
\cmidrule(lr){21-24}
Model & Head & Loss & Clust.
& $\tau_c$ & UPS & Sharp & Cov
& $\tau_c$ & UPS & Sharp & Cov
& $\tau_c$ & UPS & Sharp & Cov
& $\tau_c$ & UPS & Sharp & Cov
& $\tau_c$ & UPS & Sharp & Cov \\
\midrule
\multicolumn{24}{l}{Before Calibration} \\
\midrule
MC Dropout-B & - & - & No
& 78.2 & 40.2 & 0.23 & 68.3
& 61.5 & 57.9 & 0.15 & 62.4
& 55.3 & 39.1 & 0.25 & 65.8
& 59.4 & 38.5 & 0.17 & 64.6
& 58.9 & 15.3 & 0.30 & 46.3 \\
MC Dropout-L & - & - & No
& 78.8 & 42.1 & 0.18 & 63.2
& 63.2 & 60.8 & 0.11 & 68.4
& 56.7 & 42.9 & 0.21 & 61.0
& 60.2 & 48.0 & 0.12 & 58.8
& 59.5 & 13.7 & 0.24 & 42.6 \\
BEiTS++-B & Unc. & PinB. & No
& 71.7 & 14.4 & 0.02 & 3.5
& 58.7 & 22.2 & 0.01 & 0.3
& 52.1 & 10.8 & 0.01 & 13.9
& 45.1 & 15.6 & 0.02 & 2.9
& 59.9 & 12.4 & 0.04 & 18.5 \\
BEiTS++-B & Ch. & PinB. & No
& 78.4 & 42.9 & 0.27 & 60.1
& 61.6 & 58.7 & 0.16 & 33.9
& 56.0 & 47.5 & 0.31 & 49.8
& 59.8 & 53.1 & 0.20 & 34.9
& 58.8 & 25.5 & 0.36 & 42.2 \\
BEiTS++-B & Ch. & CI & No
& 78.5 & 43.9 & 0.28 & 61.5
& 61.4 & 59.8 & 0.16 & 34.2
& 56.1 & 48.0 & 0.31 & 50.9
& 59.9 & 54.5 & 0.21 & 36.6
& 59.0 & 25.8 & 0.37 & 43.8 \\
\midrule
BEiTS++-B & Ch. & CI & Yes
& 78.5 & 43.8 & 0.28 & 61.4
& 61.5 & 59.7 & 0.16 & 34.1
& 56.1 & 47.9 & 0.31 & 50.6
& 59.9 & 54.2 & 0.21 & 36.4
& 59.0 & 25.9 & 0.37 & 43.5 \\

BEiTS++-L & Ch. & CI & Yes
& 79.0 & 46.2 & 0.27 & 59.1
& 63.0 & 57.5 & 0.15 & 34.2
& 57.2 & 44.2 & 0.30 & 50.1
& 60.7 & 55.8 & 0.19 & 32.8
& 59.4 & 24.8 & 0.34 & 41.5 \\
\midrule
\multicolumn{24}{l}{After Calibration} \\
\midrule
MC Dropout-B & - & - & No
& 78.2 & 43.5 & 0.34 & 85.5
& 61.5 & \textbf{60.6} & 0.22 & 97.3
& 55.3 & 42.1 & \underline{0.38} & 84.7
& 59.4 & 45.8 & 0.23 & 76.8
& 58.9 & 20.0 & 0.45 & 68.7 \\
MC Dropout-L & - & - & No
& \underline{78.8} & \underline{45.6} & \textbf{0.29} & 83.1
& \textbf{63.2} & \underline{60.2} & \textbf{0.17} & 97.0
& \underline{56.7} & 43.4 & \textbf{0.34} & \underline{85.2}
& \underline{60.2} & 50.4 & \textbf{0.19} & 73.4
& \underline{59.5} & 21.5 & \textbf{0.39} & 65.0 \\
BEiTS++-B & Unc. & PinB. & No
& 71.7 & 11.8 & 0.49 & 94.5
& 58.7 & 18.3 & 0.48 & \textbf{95.4}
& 52.1 & 7.9 & 0.49 & 83.7
& 45.1 & 12.6 & 0.49 & \textbf{94.8}
& \textbf{59.9} & 10.0 & 0.51 & \textbf{81.0} \\
BEiTS++-B & Ch. & PinB. & No
& 78.4 & 42.8 & \underline{0.30} & 86.6
& 61.6 & 58.6 & \textbf{0.17} & \underline{96.7}
& 56.0 & 47.4 & \textbf{0.34} & 79.4
& 59.8 & 53.0 & \underline{0.22} & 77.6
& 58.8 & 25.5 & \underline{0.40} & 61.1 \\
BEiTS++-B & Ch. & CI & No
& 78.5 & 43.9 & 0.31 & \underline{86.9}
& 61.4 & 59.8 & \underline{0.18} & 97.0
& 56.1 & \textbf{48.0} & \textbf{0.34} & 78.9
& 59.9 & \underline{54.4} & 0.23 & 79.0
& 59.0 & \underline{25.9} & 0.41 & 59.2 \\
\midrule
BEiTS++-B & Ch. & CI & Yes
& 78.5 & 43.8 & 0.34 & \textbf{90.5}
& 61.5 & 59.5 & 0.21 & 97.6
& 56.1 & \underline{47.9} & \underline{0.38} & \textbf{85.6}
& 59.9 & 54.1 & 0.26 & \underline{82.8}
& 59.0 & \textbf{26.0} & 0.45 & \underline{69.9} \\
BEiTS++-L & Ch. & CI & Yes
& \textbf{79.0} & \textbf{46.1} & \underline{0.30} & 86.8
& \underline{63.0} & 57.4 & \underline{0.18} & 97.4
& \textbf{57.2} & 44.1 & \textbf{0.34} & 81.2
& \textbf{60.7} & \textbf{55.7} & \underline{0.22} & 77.1
& 59.4 & 24.8 & \textbf{0.39} & 63.7 \\
\bottomrule
\end{tabular}
}
\vspace{-0.12cm}
\caption{Quantitative results across multiple datasets and model configurations, before and after conformal calibration at a target coverage of 90\%. Head indicates an unconstrained (Unc.) or chained (Ch.) quantile head, Loss the pinball (PinB.) or centered interval (CI) loss, and Clust. the use of subgroup-based batching and calibration. For MC dropout, $\tau_c$ is that of the deterministic BEiTScore model. \textbf{Bold} indicates the best value and \underline{underline} the second-best value for each metric within each dataset, after calibration. For sharpness, lower values are better, and for coverage, values closer to the 90\% target are better.}
\vspace{-0.12cm}
\label{tab:uncertainty_hj}
\end{table*}

Our main goal is to assess whether the estimated uncertainty, measured by the width of the predicted confidence intervals, correlates well with the model's prediction error, defined as the absolute difference between the human rating and the predicted 50th percentile. In parallel, we measure Kendall's $\tau_c$ between the resulting median scores and the human ratings. Ideally, we want BEiTScore++ to match the state-of-the-art correlation with human judgment results of the original BEiTScore metric. We also analyze sharpness, to determine whether improved uncertainty estimation comes at the cost of wider intervals. Together, these metrics allow us to evaluate whether the model achieves a useful trade-off between calibration, sharpness, correlation with human judgments, and the extent to which higher uncertainty is associated with larger deviations from human ratings.

Table~\ref{tab:uncertainty_hj} presents a quantitative comparison across multiple datasets and model configurations, including a MC dropout baseline over the original BEiTScore, the proposed uncertainty-aware BEiTScore++ variants, and a set of ablations that justify our design choices. Specifically, the table includes: (i) MC dropout as a uncertainty baseline, evaluated with the BEiTScore backbone in both base and large variants; (ii) a variant where the quantile regression head predicts each quantile independently, without the proposed chained mechanism, trained with the standard pinball loss; (iii) a chained head trained with the pinball loss, which isolates the contribution of the loss; and (iv) a variant that does not use our subgroup-based batching strategy, nor the subgroup-wise calibration procedure. 

The chained head accounts for most of the gain over the unconstrained one, whose intervals are extremely narrow but contain the human rating in at most 18.5\% of the cases before calibration, raising UPS from 10.8--22.2 to 25.5--58.7, while the centered interval loss adds a smaller but consistent gain of up to 1.4 points. After calibration, coverage of the chained models ranges from 78.9\% to 97.6\% in domain, against the 90\% target, but falls to 59.2--82.8\% on Flickr8K-Expert and Composite, as expected from a marginal guarantee that assumes exchangeability. Clustering leaves the uncalibrated results essentially unchanged, but after calibration raises coverage on every benchmark, e.g., from 59.2\% to 69.9\% on Composite, with intervals wider by 0.03--0.04. Against Monte Carlo dropout, BEiTScore++ achieves a higher UPS on four of five benchmarks for both sizes, with Monte Carlo dropout ahead on Polaris, while requiring one forward pass rather than 25. The median-score correlation of BEiTScore++ differs from that of BEiTScore by at most 0.8 points, so uncertainty estimation does not degrade scoring.

A further consequence of omitting the chained mechanism concerns how the
intervals can be calibrated. When the quantile regression head predicts each quantile independently, we observe quantile crossing. This prevents
us from applying the same two-parameter conformal calibration of
Equations~\ref{eq:cal_lower} and~\ref{eq:cal_upper}. In
the presence of crossing the multiplicative factor becomes ill-defined, so we are forced to set $\omega$ to zero and calibrate only $\beta$, which reduces Equations~\ref{eq:cal_lower} and~\ref{eq:cal_upper} to a symmetric band around the median. Restricting the calibration to this single degree of freedom yields intervals that reach high coverage only by becoming very wide (sharpness of 0.48 to 0.51), and whose width remains weakly related to the prediction error.


\begin{table*}[t]
\centering
\small
\resizebox{0.76\linewidth}{!}{%
\begin{tabular}{l c c c c c c c c}
\toprule
\textbf{Model} &\textbf{Size (M)} & \textbf{GFLOPs} & & \textbf{VICR$\tau_c$} & \textbf{Polaris$\tau_c$} & \textbf{Nebula$\tau_c$} & \textbf{EX-8K$\tau_c$} & \textbf{COM$\tau_c$} \\
\midrule
\rowcolor{gray!15}
\multicolumn{9}{c}{\textit{LLM-based Methods}} \\
\midrule
FLEUR~\cite{lee2024fleur}   & $13350.84$ & $9317.47$ & & $74.7$ & $58.4$ & $52.7$ & $53.0$ & \underline{60.8} \\
EXPERT~\cite{kim2025expert}    & $13350.84$ & $9317.47$ & & $77.4$ & $60.9$ & $54.9$ & $56.7$ & \textbf{61.8} \\
VQAScore (CLIP-FlanT5-XXL)~\cite{lin2024evaluating} & $11459.59$ & $3742.10$ & & $69.8$ & $51.6$ & $49.4$ & $41.0$ & $53.4$ \\
\midrule
\rowcolor{gray!15}
\multicolumn{9}{c}{\textit{Encoder-based Methods}} \\
\midrule
CLIP-S/ViT-B~\cite{hessel2021clipscore} & $151.28$ & $7.28$ & & $70.8$ & $52.9$ & $47.8$ & $52.3$ & $51.8$ \\
CLIP-S/ViT-L~\cite{hessel2021clipscore} & $427.62$ & $84.36$ & & $72.6$ & $53.2$ & $48.8$ & $54.7$ &  $55.4$ \\
LongCLIP-S/ViT-B~\cite{zhang2024long}   & $149.84$ & $26.24$ & & $70.2$ & $54.8$ & $48.9$ & $53.0$ &  $51.3$ \\
LongCLIP-S/ViT-L~\cite{zhang2024long}   & $427.75$ & $98.90$ & & $69.0$ & $53.9$ & $48.6$ & $53.9$ & $50.5$ \\
PAC-S/ViT-B~\cite{sarto2023positive}    & $151.28$ & $4.88$ & & $71.4$ & $52.3$ & $47.2$ & $54.3$ & $53.5$ \\
PAC-S/ViT-L~\cite{sarto2023positive}    & $427.62$ & $56.26$ & & $73.2$ & $53.1$ & $47.9$ & $55.7$ &  $55.4$ \\
PAC-S++/ViT-B~\cite{sarto2025positive}  & $155.42$ & $4.88$ & & $72.7$ & $52.4$ & $46.9$ & $54.5$ & $55.7$ \\
PAC-S++/ViT-L~\cite{sarto2025positive}  & $430.58$ & $56.26$ & & $75.2$ & $53.6$ & $48.9$ & $57.4$ & $59.4$ \\
SPECS~\cite{chen2025specs}              & $151.49$ & $9.20$  & & $33.4$ & $39.7$ & $34.4$ & $30.4$& $24.1$ \\
\midrule
BEiTScore-B/32                & $222.28$ & $83.85$ & & $78.2$ & $61.5$ & $55.3$ & $59.4$ & $58.9$  \\
BEiTScore-L/14                & $674.69$ & $296.73$ & & \underline{78.8} & \textbf{63.2} & \underline{56.7} & \underline{60.2} & $59.5$  \\
\midrule
BEiTScore++-B/32                & $222.28$ & $83.85$ & & $78.5$ & $61.5$ & $56.1$ & $59.9$ & $59.0$  \\
BEiTScore++-L/14                & $674.69$ & $296.73$ & & \textbf{79.0} & \underline{63.0} & \textbf{57.2} & \textbf{60.7} & $59.4$  \\
\bottomrule
\end{tabular}}
\vspace{-0.12cm}
\caption{Comparison against state-of-the-art reference-free metrics across multiple human correlation benchmarks. Methods are categorized as encoder-based or LLM-based. Performance is measured using Kendall’s $\tau_c$ correlation coefficient with human judgments on all benchmarks. The highest score is highlighted using \textbf{bold}, and the second best is highlighted using an \underline{underline}.}
\vspace{-0.12cm}
\label{tab:human_corr}
\end{table*}
In complement to the results presented in Table~\ref{tab:uncertainty_hj}, Table~\ref{tab:human_corr} provides a more detailed evaluation of the correlation between the median predicted scores and human judgments, across a wide range of benchmarks. We compare the proposed uncertainty-aware BEiTScore++ metric against several representative reference-free metrics from the literature, including both encoder-based and LLM-based approaches. This analysis is intended to verify that introducing uncertainty estimation does not compromise the model’s ability to produce accurate quality assessments.

Overall, the results confirm that our uncertainty-aware extension retains the performance of BEiTScore in terms of agreement with human judgments. The large version of the proposed model achieves the best result on VICR, Nebula, and Flickr8K-Expert, and the second best on Polaris, while on Composite it is surpassed by the LLM-based EXPERT and FLEUR metrics, as was already the case for BEiTScore.

\vspace{-0.13cm}
\subsection{Different Sources of Uncertainty} 
\vspace{-0.13cm}
\label{sec:different_sources_of_uncertainty}

This section investigates whether the uncertainty estimates produced by BEiTScore++ meaningfully reflect different underlying sources of uncertainty. We focus on two complementary factors: input ambiguity, arising from degradation or corruption of the input images, and model knowledge gaps, stemming from exposure to out-of-domain data. Through controlled experiments, we assess whether the uncertainty estimator responds in a predictable and interpretable manner. Results are reported for the large variant, although the base-sized model yields similar conclusions, with results being provided in Appendix~\ref{app:beitscore-b}.

\vspace{-0.13cm}
\subsubsection{Input Ambiguity}
\vspace{-0.13cm}

To examine whether BEiTScore++ responds appropriately to input ambiguity, we designed an experiment in which images from the Polaris test split were subjected to varying levels of Gaussian blur. The underlying hypothesis is that blur degrades the visual information available to the model, and consequently the model uncertainty should increase with the level of blur, at least while the image content remains recognizable.

A key preprocessing step was taken to avoid confounding effects arising from the score distribution. The original Polaris test split is heavily skewed towards low scores, with a large proportion of instances receiving a human rating of zero. For already poor image-caption pairs, the addition of blur could plausibly reinforce the model's confidence in assigning low scores, rather than inducing uncertainty through wider intervals. This behavior would obscure the relationship we aim to measure. To mitigate this, we downsampled instances with a human score of zero, and built a subset of $3624$ instances with a more uniform distribution across the full spectrum of scores, aiming for the effect of blur on model uncertainty to be observed clearly across instances of varying quality.

\begin{figure}[t]
    \centering
    \includegraphics[width=\columnwidth]{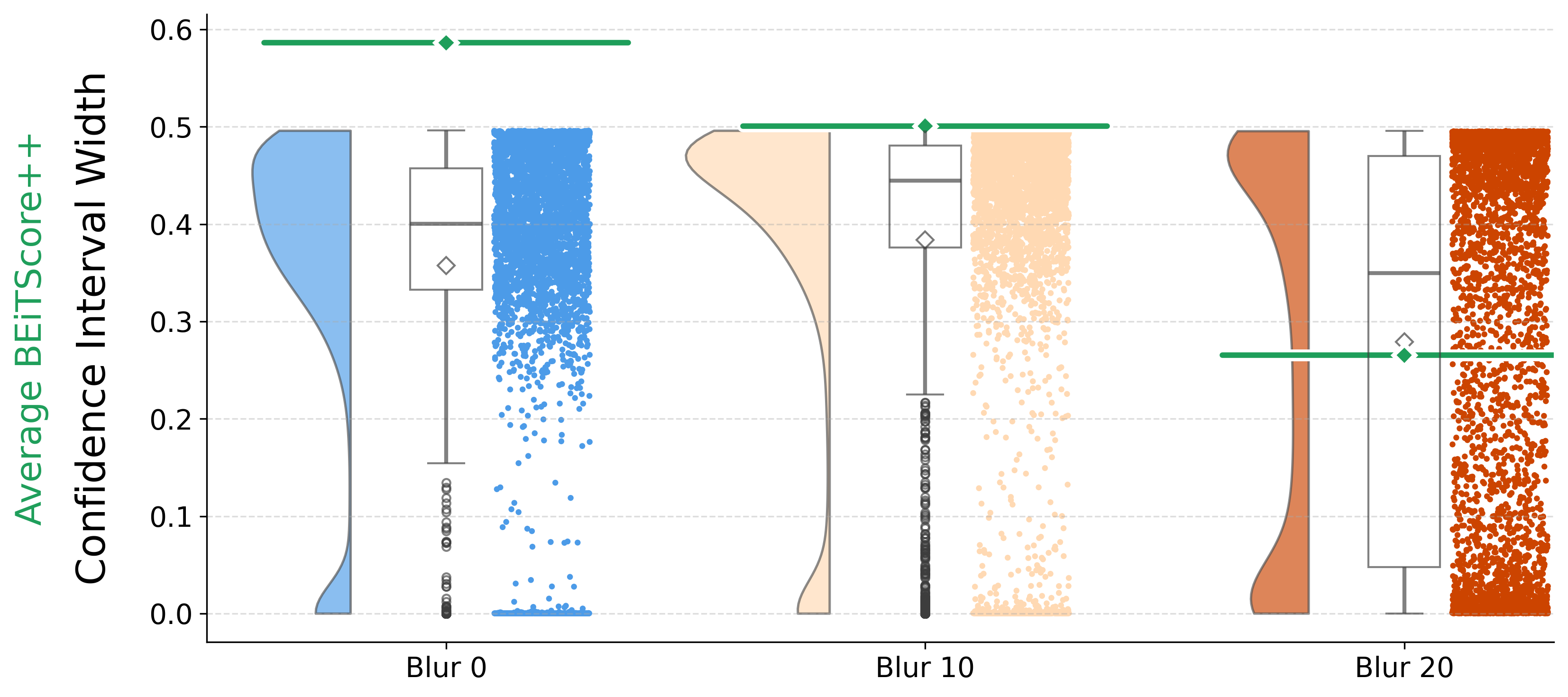}
    \vspace{-0.3cm}    
    \caption{Distribution of 90\% confidence interval widths generated by the large variant of BEiTScore++, as Gaussian blur increases on images from the Polaris test split. The horizontal lines indicate the average median score produced by BEiTScore++ at each blur level.}
\vspace{-0.13cm}    
    \label{fig:box_plot_input_amb}
\end{figure}

With the balanced subset, we applied progressively stronger levels of blur to each image and measured the corresponding model uncertainty at each level. Figure~\ref{fig:box_plot_input_amb} shows BEiTScore++ interval width distributions under different blur perturbation levels. At blur level 10, the distribution of interval widths shifts upwards relative to the original images, with the median width rising from about 0.40 to 0.45, as assessed visually from the distributions rather than through a formal hypothesis test. At blur level 20, the interval widths decrease instead, with a much larger spread. We attribute this to the images becoming so severely distorted that the model grows confident that the image and caption are misaligned, effectively starting to collapse uncertainty in favor of a low-score prediction. In addition to interval width distribution, Figure~\ref{fig:box_plot_input_amb} also presents the average BEiTScore++ ratings for each blur level, confirming that the model produces lower scores for increased input degradation.

We additionally evaluated model behavior under a broader set of image distortions, including salt-and-pepper noise, JPEG compression artifacts, random patch occlusions, elastic warping (via spatial smoothing and edge curling), and contrast adjustments (see Appendix~\ref{app:beyond_blur} for full details). For noise, compression, occlusion, and warping, the interval widths are larger than for the original images, although the increase in the mean width is modest, of about 0.01 to 0.03, whereas contrast adjustments leave them unchanged. This supports the hypothesis that the estimator is sensitive to structural degradation of the input, while being robust to photometric changes.


\vspace{-0.13cm}
\subsubsection{Model Knowledge Gaps}
\vspace{-0.13cm}

To study uncertainty arising from model knowledge gaps, we analyze the model’s confidence on in-domain datasets with natural images, including long-caption datasets IIW~\cite{garg2024imageinwords}, DOCCI~\cite{onoe2024docci}, and Localized Narratives~\cite{pont2020connecting}, as well as the short-caption dataset MSCOCO~\cite{lin2014microsoft}, and compare them against artwork datasets Docent~\cite{ananthram2025posh} and ArtCaps~\cite{lu2022artcap}. These artwork datasets are not only out-of-domain but also more ambiguous: their images are paintings rather than realistic photographs, with stylized shapes, smoother edges, and greater abstraction, while the captions, especially the detailed descriptions in Docent, often involve a higher degree of subjective interpretation.

Since BEiTScore++ supports both detailed and shorter captions, we divide the evaluation into two complementary groups: long captions, in which we compare IIW, DOCCI, and Localized Narratives against Docent; short captions, in which we compare ArtCaps against MSCOCO.

\begin{figure}[t]
    \centering
    \includegraphics[width=\columnwidth]{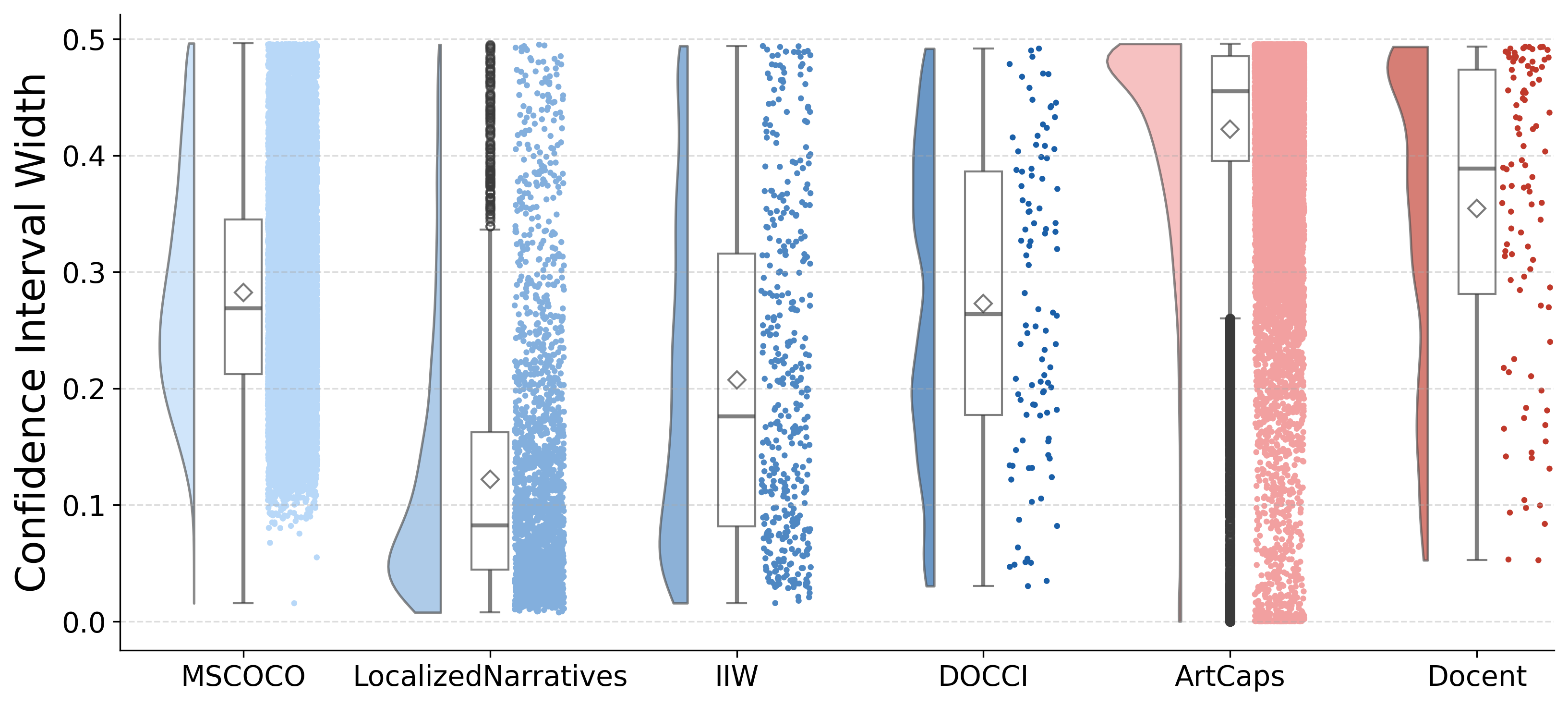}
    \vspace{-0.3cm}
    \caption{Distribution of 90\% confidence interval widths produced by the large variant of the BEiTScore++ metric across in-domain images and out-of-domain artwork datasets, and also across short- and long-caption settings. The color gradient indicates the caption length, with pastel tones for short captions, and more saturated tones for detailed descriptions.}
    \vspace{-0.13cm}
    \label{fig:model_gap}
\end{figure}

Figure~\ref{fig:model_gap} indicates that in both the short- and long-caption settings, the model exhibits higher uncertainty when evaluating artwork datasets. This is consistent with an increased epistemic uncertainty, as the model was not trained on artistic imagery, although part of the increase may reflect the greater ambiguity of artwork, which this experiment cannot separate.

\vspace{-0.13cm}
\section{Conclusions}
\vspace{-0.13cm}

We introduced an uncertainty-aware extension of BEiTScore that predicts caption quality together with a confidence interval. By combining quantile regression with a monotone construction of quantiles, together with a centered interval loss, BEiTScore++ produces coherent intervals in a single forward pass. Post-hoc conformal calibration raises coverage towards the target, and the subgroup-based rule raises it further, mainly by selecting wider intervals.

Empirical results show that our method maintains strong agreement with human judgments, while producing uncertainty estimates that correlate with prediction error and grow under input degradation and domain shift, indicating that the model captures meaningful signals about its own reliability. These properties are essential for downstream applications, where evaluation metrics inform benchmarking or model selection.

\section*{Limitations and Ethical Considerations}

While BEiTScore++ improves uncertainty estimation for image-caption evaluation, providing calibrated confidence intervals in a single forward pass, several important limitations remain. 

First, our experiments are restricted to English captions, so the behavior of the method in multilingual settings is still unknown. Extending BEiTScore++ to other languages is an important direction for future work, especially given recent evidence that CLIP-based metrics can be adapted for multilingual caption evaluation~\cite{gomes2025evaluation}. It will be important to study whether the proposed uncertainty estimates remain reliable when captions contain translation artifacts, language-specific constructions, or morphological complexity that differs from English.

Second, although our method improves evaluation interpretability, it does not remove the need for human evaluation. Automated metrics can support and complement human judgment, but they should not be treated as a substitute, particularly in applications where erroneous assessments may have downstream consequences. More generally, uncertainty intervals may help users identify low-confidence cases, but they do not guarantee fairness or correctness in all settings.

Third, the proposed framework is evaluated on benchmark datasets that may not fully capture the diversity of real-world image-caption pairs. As with many evaluation methods, its performance may be affected by dataset bias, domain shift, or limited coverage of rare concepts and styles. Future work should investigate how BEiTScore++ behaves under broader distributional shifts, and whether additional calibration strategies are needed to maintain robustness in more challenging settings.

Finally, while BEiTScore++ is designed to be efficient at inference time, the broader research area still depends on substantial computational resources for training and evaluation. In that sense, developing uncertainty-aware metrics that remain accurate, efficient, and broadly applicable across languages and domains remains as a particularly important research direction.

We also note that we used GitHub Copilot during the development of our research work, and we used ChatGPT for minor verifications during the preparation of this manuscript.

\section*{Acknowledgements}
We thank the anonymous reviewers for their valuable comments and suggestions.
This research was supported by the Portuguese Recovery and Resilience Plan through project C645008882-00000055 (i.e., the Center For Responsible AI), and by Fundação para a Ciência e Tecnologia (FCT) through the PhD scholarship with reference 2025.06139.BD, and through the projects with references 2024.07385.IACDC and PEST UID/50021/2025 (DOI: https://doi.org/10.54499/UID/50021/2025). In addition, C. Zerva was partly supported through the framework of the Pharos AI Factory project, funded by the European High-Performance Computing Joint Undertaking (EuroHPC JU) under Grant Agreement No. 101234269 as part of the Horizon Europe, and by the Greek Public Investments Program programme, as well as by a Google Research Scholar award.

\bibliography{custom}

\appendix

\section{Datasets}
\vspace{-0.13cm}

\label{app:datasets}

The following datasets were used in the training, calibration, and evaluation of BEiTScore++, specifically in experiments that focused on assessing the Uncertainty Pearson Score (UPS) and the correlation with human judgments. We separate the datasets into those used for training and calibration, and those used only for evaluation.

\paragraph{Datasets used for training and calibration.}
\begin{itemize}\setlength\itemsep{0em}
\item \textbf{VICR}~\cite{narins2023validated}:
The Validated Image Caption Rating (VICR) dataset features human ratings collected through a gamified approach for gathering image-caption pairs, involving a total of 9,990 distinct images. The authors of the dataset demonstrated that it exhibits a superior inter-rater agreement compared to other alternatives (e.g., an improvement of 19\% in Fleiss' $\kappa$ when compared to the agreement for the Flickr8K-Expert dataset), and it features a more balanced distribution across various levels of caption quality. The dataset is organized into 10,175 training pairs, 2,310 validation pairs, and 3,161 test pairs, with an average of 4.4 annotations per instance. In our experiments, we used the test split of the VICR dataset, which includes 3,161 unique image-caption pairs. For calibration, we used the original validation split, which comprises 2,310 unique image-caption pairs.
\item \textbf{Polaris}~\cite{wada2024polos}:
The Polaris dataset comprises human judgments on image-caption pairs involving 13,691 distinct images, collected from evaluators. The dataset is organized into 26,340 training pairs, 8,738 validation pairs, and 8,726 test pairs, with an average of 3.0 annotations per instance. Polaris includes captions generated by ten standard image captioning models, covering both modern and older architectures to ensure output diversity. In our tests, we used the test split of the Polaris dataset, which includes 8,726 unique image-caption pairs. For calibration, we used the validation split, which comprises 8,738 unique image-caption pairs.
\item \textbf{Nebula}~\cite{matsuda2024deneb}:
The Nebula dataset comprises 32,978 human annotations on image-caption pairs, spanning 32,978 unique images. It is organized into 26,382 training pairs, 3,298 validation pairs, and 3,298 test pairs. The dataset was introduced for training hallucination-robust image captioning evaluation metrics and was constructed by extending the Polaris dataset to increase visual diversity. In our experiments, we used the test split of Nebula, which contains 3,298 image-caption pairs. For calibration, we used the validation split, which also comprises 3,298 image-caption pairs.
\item \textbf{EvalMuse-40K}~\cite{han2024evalmuse}:
The EvalMuse-40K dataset comprises 32,717 human annotations on image-caption pairs, spanning 32,717 unique images. It is organized into 29,445 training pairs and 3,272 validation pairs. EvalMuse-40K was introduced as a fine-grained benchmark for text-to-image generation evaluation, with comprehensive human annotations designed to better assess image-text alignment. Our experiments used the training pairs for model training, and the validation pairs for calibration.
\end{itemize}

\paragraph{Datasets used only for evaluation.}
\begin{itemize}\setlength\itemsep{0em}
\item \textbf{Flickr8K-Expert}~\cite{hodosh2013framing}:
This dataset comprises 16,992 expert human judgments for 5,664 image–caption pairs from the Flickr8K dataset. Human assessors graded captions on a scale of 1 to 4, where 4 indicates a caption that accurately describes the image without errors, and 1 signifies a caption unrelated to the image.

\item \textbf{Composite}~\cite{aditya2015images}:
This dataset contains 13,146 image–caption pairs taken from MS-COCO (2,007 images), Flickr8K (997 images), and Flickr30K (991 images). Each image originally had five reference captions. One of these references was chosen for human rating and subsequently removed from the reference set that is to be used when assessing evaluation metrics.
\end{itemize}
Table~\ref{tab:dataset_statistics} summarizes the key statistics of the datasets used for training, calibration, and evaluation.
\begin{table*}[t]
\centering
\small
\setlength{\tabcolsep}{6pt}
\renewcommand{\arraystretch}{1.2}
\resizebox{\linewidth}{!}{
\begin{tabular}{lcccccc}
\toprule
Metric & VICR & Polaris & Nebula & Flickr8K-Expert & Composite & EvalMuse-40K \\
\midrule
Training & 10175 & 26340 & 26382 & - & - & 29445 \\
Validation & 2310 & 8738 & 3298 & - & - & 3272 \\
Testing & 3161 & 8726 & 3298 & 5664 & 13146 & - \\
\midrule
Unique Images & 9990 & 13691 & 32978 & 1000 & 3995 & 32717 \\
\midrule
Caption Len. $\mu \pm \sigma$ & 10.9 $\pm$ 3.0 & 9.3 $\pm$ 5.5 & 8.8 $\pm$ 2.5 & 11.9 $\pm$ 4.2 & 13.1 $\pm$ 9.3 & 11.0 $\pm$ 7.0 \\
(Q1/Q2/Q3) & \small (9.0/10.0/12.0) & \small (8.0/10.0/11.0) & \small (8.0/9.0/10.0) & \small (9.0/11.0/14.0) & \small (9.0/11.0/14.0) & \small (6.0/9.0/14.0) \\
\midrule
Annot./Inst. $\mu \pm \sigma$ & 4.4 $\pm$ 0.8 & 3.0 $\pm$ 2.8 & 1.0 $\pm$ 0.0 & 3.0 $\pm$ 0.0 & 1.0 $\pm$ 0.0 & 1.0 $\pm$ 0.0 \\
(Q1/Q2/Q3) & \small (4.0/4.0/5.0) & \small (1.0/1.0/6.0) & \small (1.0/1.0/1.0) & \small (3.0/3.0/3.0) & \small (1.0/1.0/1.0) & \small (1.0/1.0/1.0) \\
\midrule
Human Score $\mu \pm \sigma$ & 2.67 $\pm$ 1.41 & 0.25 $\pm$ 0.34 & 0.52 $\pm$ 0.36 & 1.58 $\pm$ 0.72 & 3.00 $\pm$ 1.45 & 0.46 $\pm$ 0.25 \\
(Q1/Q2/Q3) & \small (1.20/2.40/4.19) & \small (0.00/0.00/0.62) & \small (0.00/0.60/0.78) & \small (1.00/1.33/2.00) & \small (1.50/3.00/4.50) & \small (0.25/0.42/0.67) \\
\midrule
Rating Scale & [1.0, 5.0] & [0.0, 1.0] & [0.0, 1.0] & [1.0, 4.0] & [1.0, 5.0] & [0.0, 1.0] \\[1.5em]
Score Distribution & \begin{sparkline}{10} \definecolor{sparkbottomlinecolor}{gray}{0.8} \setlength\sparkbottomlinethickness{0.7pt} \sparkbottomlinex -0.05 1.05 \sparkspike 0.0000 3.0000 \sparkspike 0.1111 0.9774 \sparkspike 0.2222 0.8193 \sparkspike 0.3333 0.7337 \sparkspike 0.4444 0.4764 \sparkspike 0.5556 0.6851 \sparkspike 0.6667 0.7269 \sparkspike 0.7778 0.6126 \sparkspike 0.8889 1.6099 \sparkspike 1.0000 1.0678 \end{sparkline} & \begin{sparkline}{10} \definecolor{sparkbottomlinecolor}{gray}{0.8} \setlength\sparkbottomlinethickness{0.7pt} \sparkbottomlinex -0.05 1.05 \sparkspike 0.0000 3.0000 \sparkspike 0.1111 0.0113 \sparkspike 0.2222 0.0311 \sparkspike 0.3333 0.0475 \sparkspike 0.4444 0.0960 \sparkspike 0.5556 0.3494 \sparkspike 0.6667 0.3836 \sparkspike 0.7778 0.4910 \sparkspike 0.8889 0.2959 \sparkspike 1.0000 0.0559 \end{sparkline} & \begin{sparkline}{10} \definecolor{sparkbottomlinecolor}{gray}{0.8} \setlength\sparkbottomlinethickness{0.7pt} \sparkbottomlinex -0.05 1.05 \sparkspike 0.0000 3.0000 \sparkspike 0.1111 0.0266 \sparkspike 0.2222 0.3205 \sparkspike 0.3333 0.1195 \sparkspike 0.4444 0.2461 \sparkspike 0.5556 1.8675 \sparkspike 0.6667 1.0043 \sparkspike 0.7778 2.0169 \sparkspike 0.8889 0.7750 \sparkspike 1.0000 1.7261 \end{sparkline} & \begin{sparkline}{10} \definecolor{sparkbottomlinecolor}{gray}{0.8} \setlength\sparkbottomlinethickness{0.7pt} \sparkbottomlinex -0.05 1.05 \sparkspike 0.0000 3.0000 \sparkspike 0.1111 1.2319 \sparkspike 0.2222 0.8796 \sparkspike 0.3333 0.8655 \sparkspike 0.4444 0.4136 \sparkspike 0.5556 0.2464 \sparkspike 0.6667 0.3064 \sparkspike 0.7778 0.0932 \sparkspike 0.8889 0.0626 \sparkspike 1.0000 0.1315 \end{sparkline} & \begin{sparkline}{10} \definecolor{sparkbottomlinecolor}{gray}{0.8} \setlength\sparkbottomlinethickness{0.7pt} \sparkbottomlinex -0.05 1.05 \sparkspike 0.0000 2.9532 \sparkspike 0.1111 1.5783 \sparkspike 0.2222 1.6896 \sparkspike 0.3333 1.3989 \sparkspike 0.4444 0.0455 \sparkspike 0.5556 1.5379 \sparkspike 0.6667 1.2447 \sparkspike 0.7778 1.4722 \sparkspike 0.8889 1.6921 \sparkspike 1.0000 3.0000 \end{sparkline} & \begin{sparkline}{10} \definecolor{sparkbottomlinecolor}{gray}{0.8} \setlength\sparkbottomlinethickness{0.7pt} \sparkbottomlinex -0.05 1.05 \sparkspike 0.0000 1.5581 \sparkspike 0.1111 1.5781 \sparkspike 0.2222 2.1917 \sparkspike 0.3333 2.3231 \sparkspike 0.4444 2.0014 \sparkspike 0.5556 3.0000 \sparkspike 0.6667 1.4776 \sparkspike 0.7778 1.5862 \sparkspike 0.8889 0.9904 \sparkspike 1.0000 0.9813 \end{sparkline} \\
\bottomrule
\end{tabular}
}
\caption{Statistical characterization of the datasets. For each metric (i.e., Caption Length, Annotations per Instance, and Human Score), we report the mean $\mu$ and standard deviation $\sigma$, along with quartiles (Q1: 25th percentile, Q2: median, Q3: 75th percentile). Score distributions are represented using sparkline histograms computed over the average human scores using the observed annotation scale inferred directly from each dataset.}
\label{tab:dataset_statistics}
\end{table*}
\newpage
\noindent \textbf{Additional datasets.}
\vspace{0.13cm}

The following datasets were used in our experiments to evaluate the proposed uncertainty estimator under different conditions, including controlled input ambiguity and model knowledge gaps. Specifically, we rely on them to assess calibration, sharpness, and correlation with human judgments, as well as to study how uncertainty changes under out-of-domain settings and varying caption lengths. IIW, DOCCI, Localized Narratives, and Docent are all datasets with long, detailed captions, making them particularly suitable for analyzing the behavior of the model in the long-caption setting.

\textbf{IIW}~\cite{garg2024imageinwords}:
ImageInWords (IIW) is a human-in-the-loop dataset designed to support rich, detailed image descriptions. It was created to capture hyper-detailed visual semantics and to facilitate the evaluation of models that generate long-form captions with fine-grained content. In our experiments, we use IIW as an in-domain dataset with realistic natural imagery and detailed textual descriptions, making it suitable for studying model behavior under standard captioning conditions. We evaluate results on the complete set of 400 test instances provided by the dataset.

\textbf{DOCCI}~\cite{onoe2024docci}:
Descriptions of Connected and Contrasting Images (DOCCI) is a dataset of long, human-authored descriptions for approximately 15k images. The annotations are designed to be comprehensive and compositional, with an emphasis on challenges such as spatial relations, counting, text rendering, and world knowledge. We use DOCCI as an in-domain benchmark for long-caption evaluation, since it provides natural images paired with detailed descriptions that stress the model’s ability to assess rich semantic alignment. In our experiments, we use the 100 test instances available in the dataset.

\textbf{Localized Narratives}~\cite{pont2020connecting}:
Localized Narratives is a large-scale multimodal dataset in which annotators describe images while simultaneously hovering over the regions they are referring to, producing word-level grounding information. The dataset covers hundreds of thousands of images from sources such as COCO, Flickr30k, ADE20K, and Open Images, and is widely used for dense vision-language understanding. In our experiments, we use it as an in-domain dataset because it contains realistic scenes and natural language descriptions that are well aligned with standard image captioning evaluation settings. Since the test split contains more than 100k instances, we randomly sample 3k examples for our analysis.

\textbf{MSCOCO}~\cite{lin2014microsoft}:
The Microsoft COCO Captions dataset is one of the most widely used benchmarks for image captioning, containing multiple human-generated captions for each image. It provides natural photographs with diverse everyday scenes and serves as a strong in-domain reference for both caption generation and evaluation. We use MSCOCO as an in-domain dataset in our experiments, including a random subset matched in size to ArtCaps for the short-caption setting. Specifically, we randomly sample 18k MSCOCO instances to compare directly against ArtCaps.

\textbf{Docent}~\cite{ananthram2025posh}:
Docent is an artwork-oriented dataset used to evaluate captioning models under domain shift. Unlike natural-image benchmarks, it focuses on artistic imagery, where semantic interpretation can be more abstract and stylistically driven. We use Docent as an out-of-domain dataset to test whether the proposed uncertainty estimator reflects model knowledge gaps when processing artwork rather than photographic content. The dataset contains 100 distinct image-caption pairs, and we use the provided reference captions for our analysis.

\textbf{ArtCaps}~\cite{lu2022artcap}:
ArtCaps is a fine-art painting captioning dataset containing paintings paired with multiple human-written descriptions. It was constructed specifically for image captioning in the artistic domain, where visual content is often complex and culturally contextual. We use ArtCaps as an out-of-domain benchmark in our experiments, since it provides a clear contrast to natural-image datasets and is well suited for probing epistemic uncertainty under domain shift. The dataset contains approximately 18k instances, which we compare against a randomly sampled subset of 18k MSCOCO instances in the short-caption setting.

\section{Sensitivity to the Number of Calibration Clusters}
\label{app:cluster-sweep}

We recall that the value $k = 63$ reported in the paper was not tuned on any coverage or width criterion.
It was obtained from the elbow and silhouette curves of the $k$-means fit on the calibration image embeddings, that is, from an unsupervised criterion computed before any conformal procedure is run.
This appendix reports a sensitivity analysis of the coverage and mean interval width of BEiTScore++-L on the different test sets, across different degrees of clustering.

For each value of $k$ we cluster the validation split of the merged EvalMuse, Nebula, Polaris and VICR data, run the conformal prediction procedure at a target coverage of $90\%$ inside every cluster to obtain a scaling factor $\omega$ and an additive bias $\beta$, and retain the most conservative pair across clusters.
Those parameters are applied to the original uncalibrated quantiles of each benchmark, and we report the resulting empirical coverage and mean interval width.
The point $k = 1$ is the unclustered calibration and serves as a baseline.

\begin{figure}[htbp]
    \centering
    \includegraphics[width=\columnwidth]{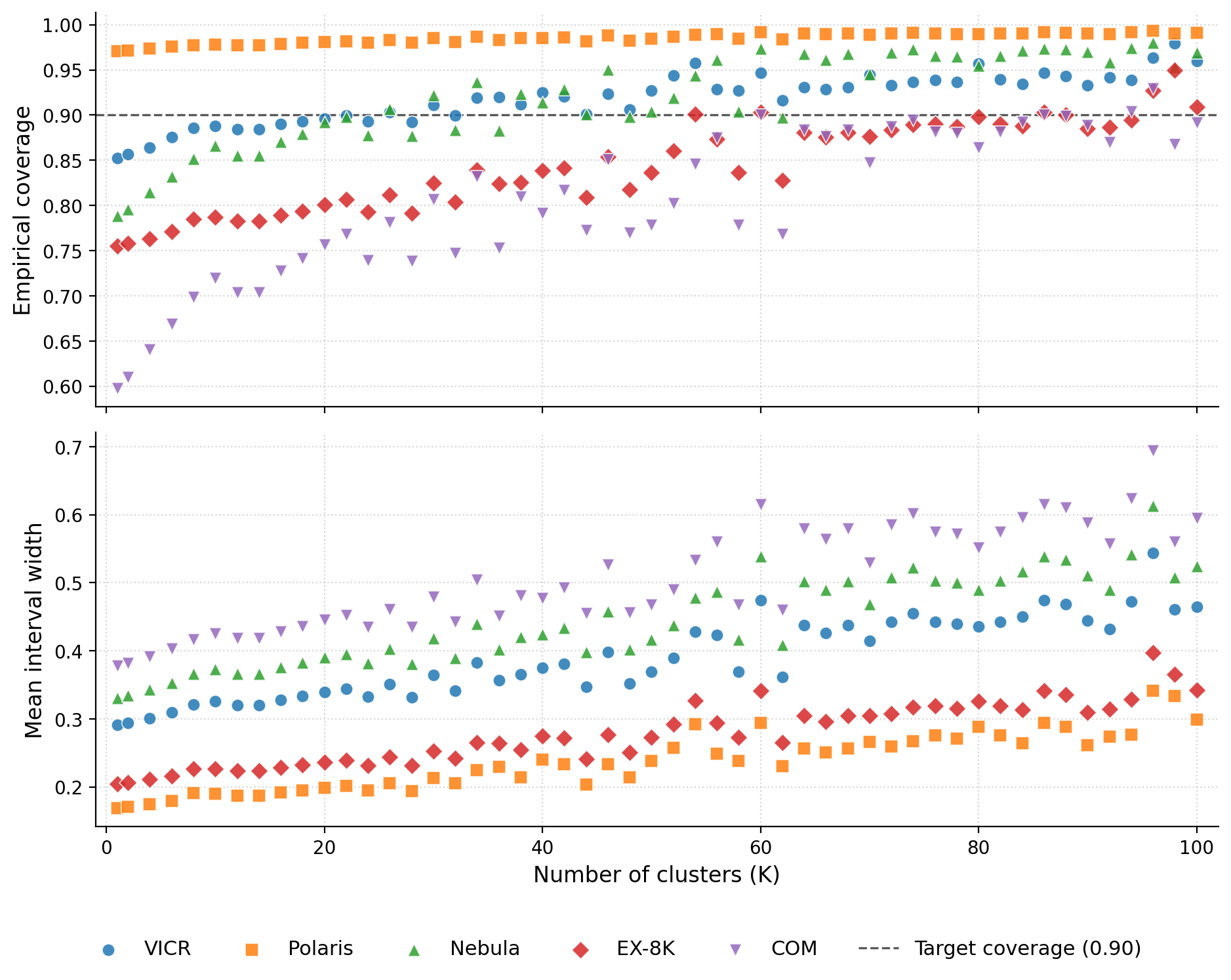}
    \caption{
        Empirical coverage (top) and mean interval width (bottom) of BEiTScore++-L as a function of the number of calibration clusters $k$.
        Colors identify the benchmark and are shared between the two panels.
        The dashed line marks the target coverage of $0.90$.
        The point at $k = 1$ corresponds to calibrating on the whole validation split without clustering.
    }
    \label{fig:cluster-sweep}
\end{figure}

Figure~\ref{fig:cluster-sweep} shows that coverage and width move together with $k$. The parameters applied at test time are the most conservative pair found across clusters, so increasing $k$ increases the number of candidates over which that maximum is taken, and the widening is attributable to the selection rule rather than to a finer conditioning on image content. The number of clusters therefore acts mainly as a knob that trades sharpness for coverage.

\section{Input Ambiguity Beyond Blur}
\label{app:beyond_blur}
\begin{figure*}[t]
    \centering
    \includegraphics[width=\linewidth]{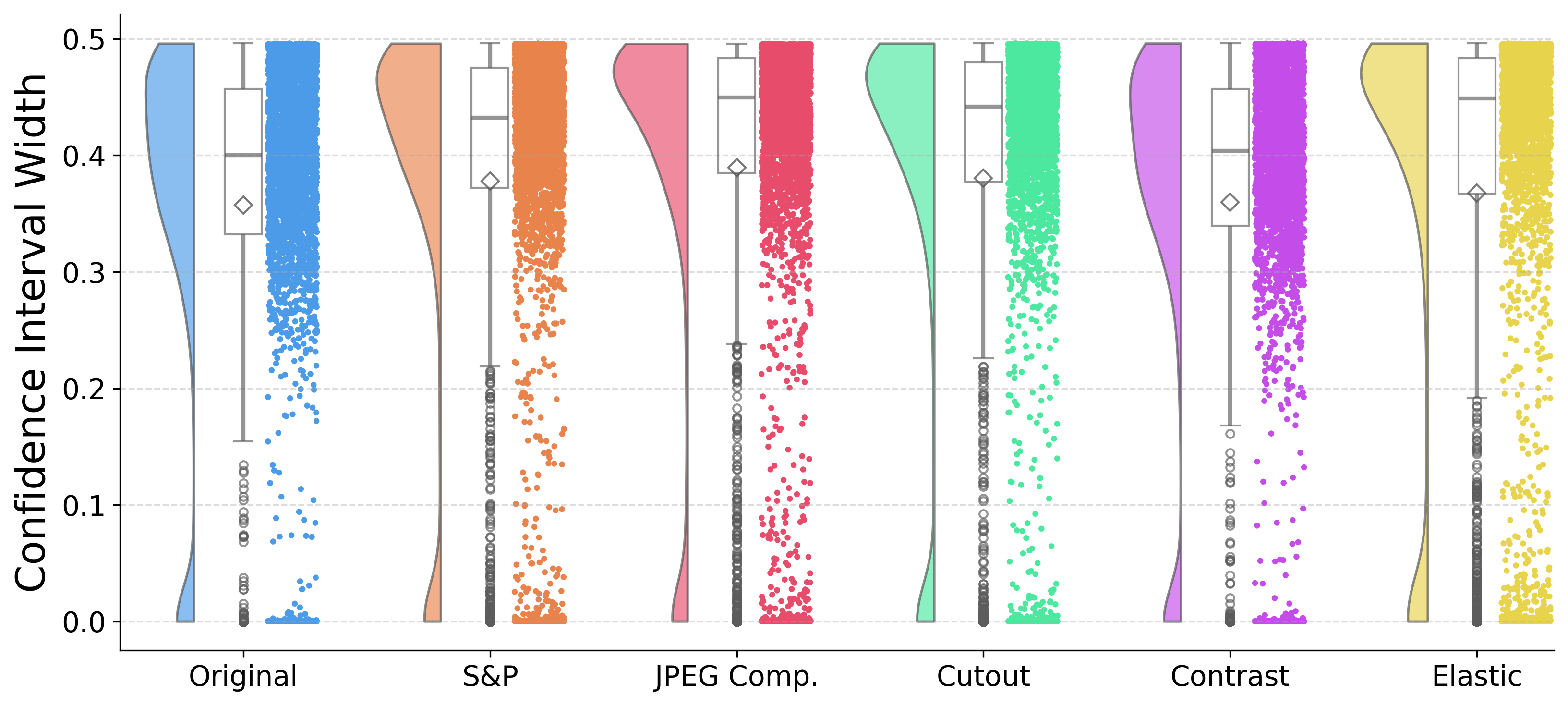}
    \caption{Box plots of confidence interval widths produced by the uncertainty estimator of BEiTScore++-L, under different types of image distortions.}
    \label{fig:box_plot_distortions}
\end{figure*}

\begin{figure*}[t]
    \centering
    \includegraphics[width=\linewidth]{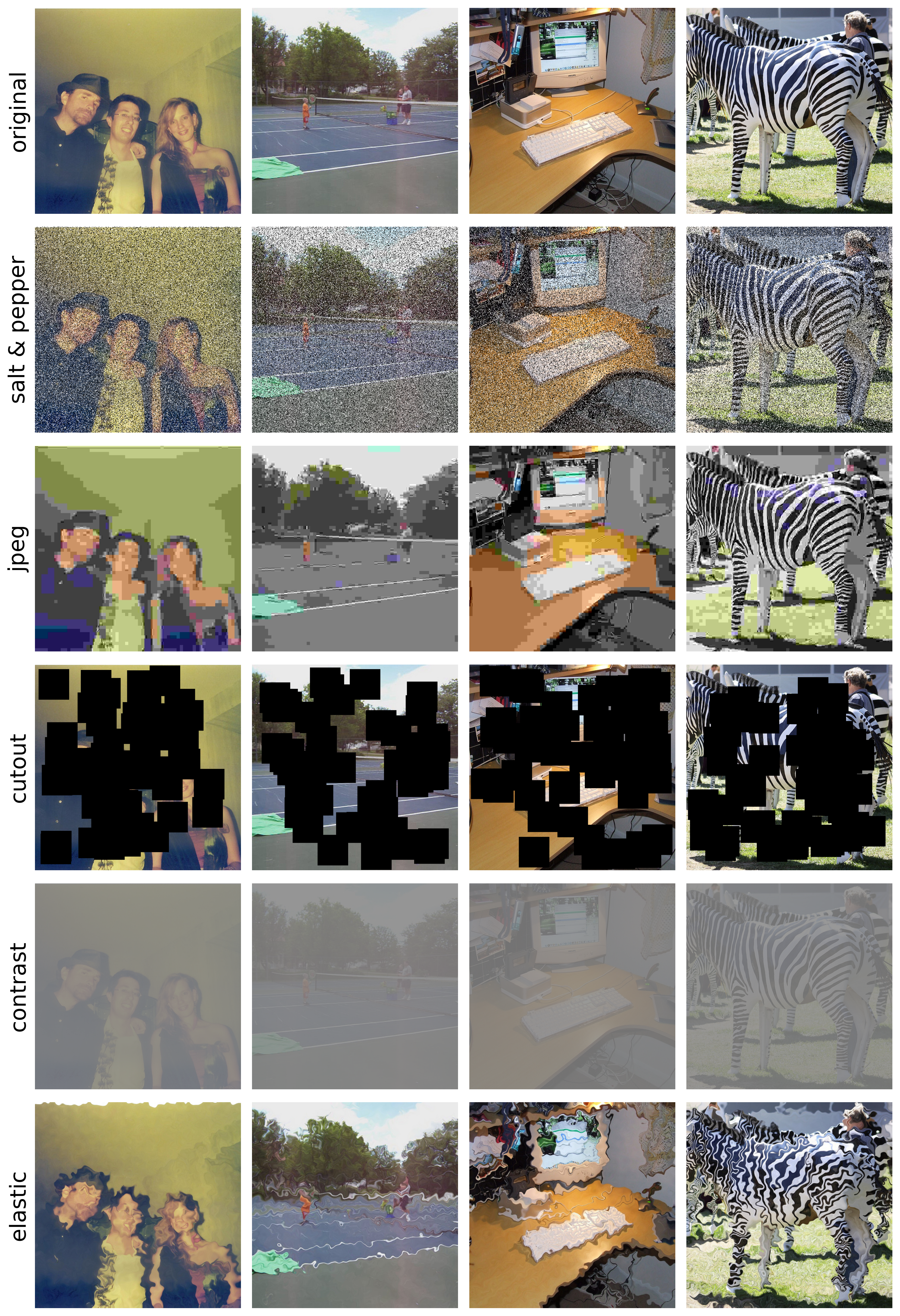}
    \caption{Examples of image distortions used to evaluate input ambiguity beyond Gaussian blur. Each row corresponds to a different type of perturbation.}
    \label{fig:distortion_grid}
\end{figure*}

While Gaussian blur provides a controlled way to degrade visual information, it
represents only one form of input ambiguity. In practice, images may be affected
by a wide range of distortions that alter different aspects of the visual signal,
such as noise, compression artifacts, structural occlusions, or geometric
deformations. To assess whether our conclusions for BEiTScore++ generalize beyond
blur, we extend the analysis to additional perturbations, including
salt-and-pepper noise, JPEG compression, random patch cutout, contrast
adjustments, and elastic warping. Figure~\ref{fig:distortion_grid} presents a
grid of examples illustrating these different distortions.

We apply each distortion to the images and measure the corresponding uncertainty
estimates. The resulting distributions of confidence interval widths are
presented in Figure~\ref{fig:box_plot_distortions}. For most distortion types, we
observe that introducing degradation leads to a modest increase in uncertainty compared to
the no-distortion setting. This trend is consistent with the results obtained
using Gaussian blur, and further supports the hypothesis that the proposed
estimator is sensitive to input ambiguity.

An exception is observed for contrast adjustments, where the interval widths
remain unchanged. We find similar behavior when applying other color-based
transformations, suggesting that the model is relatively robust to photometric
changes that preserve the underlying geometry and shape of the image content. In
contrast, distortions that disrupt structural or spatial information (e.g., noise,
occlusions, and warping) lead to an increase in model uncertainty.

\section{Analysis of Different Sources of Uncertainty Beyond BEiTScore++-L}
\label{app:beitscore-b}

In this appendix, we report the results obtained with the base-sized version of
BEiTScore++ for the same set of experiments described in
Section~\ref{sec:different_sources_of_uncertainty}. Overall, the base model
exhibits the same behavior as the large model, confirming that the observed
trends are not specific to model scale. In particular, the uncertainty estimates
increase with moderate input corruptions, remain stable under
color-based perturbations, and rise on out-of-domain artwork datasets, in
comparison to the in-domain natural image benchmarks.

\subsection{Input Ambiguity}

\begin{figure*}[t]
    \centering
    \includegraphics[width=\linewidth]{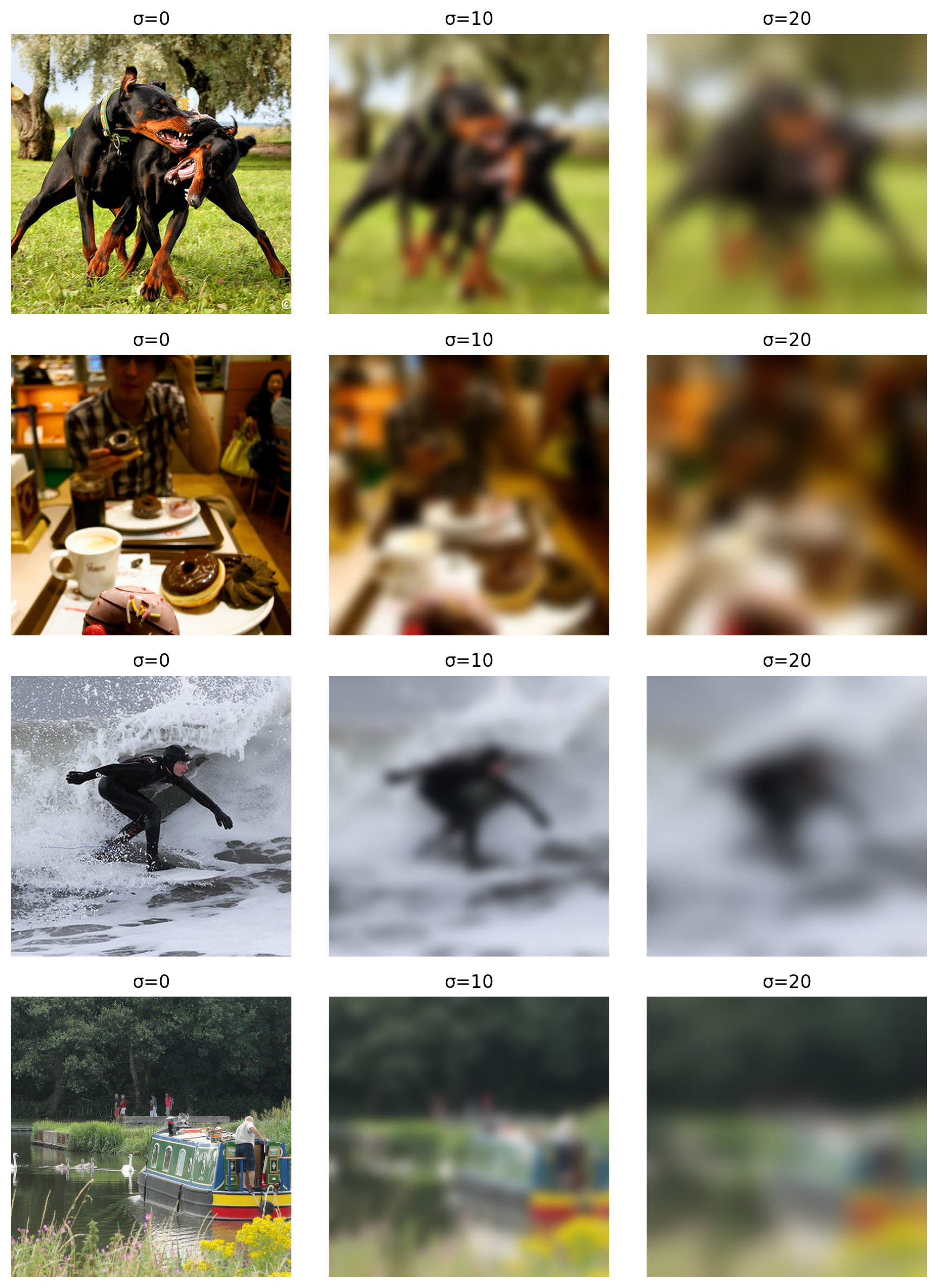}
    \caption{Examples showing different levels of blur applied to four images from the Polaris test split, used during the evaluation of input ambiguity.}
    \label{fig:blur_examples}
\end{figure*}

We first analyze the effect of input ambiguity under Gaussian blur. Some visual
examples of the instance images, with different blur levels applied to them, can
be seen in Figure~\ref{fig:blur_examples}. We use the Polaris test split for this analysis and progressively increase the blur strength applied to each image. To reduce the confounding effect of the highly skewed score distribution in Polaris, we again downsample instances with a human score of zero, obtaining a more balanced set of image-caption pairs. This allows us to more clearly examine how uncertainty changes as the visual information becomes progressively degraded.

\begin{figure*}[t]
    \centering
    \includegraphics[width=\linewidth]{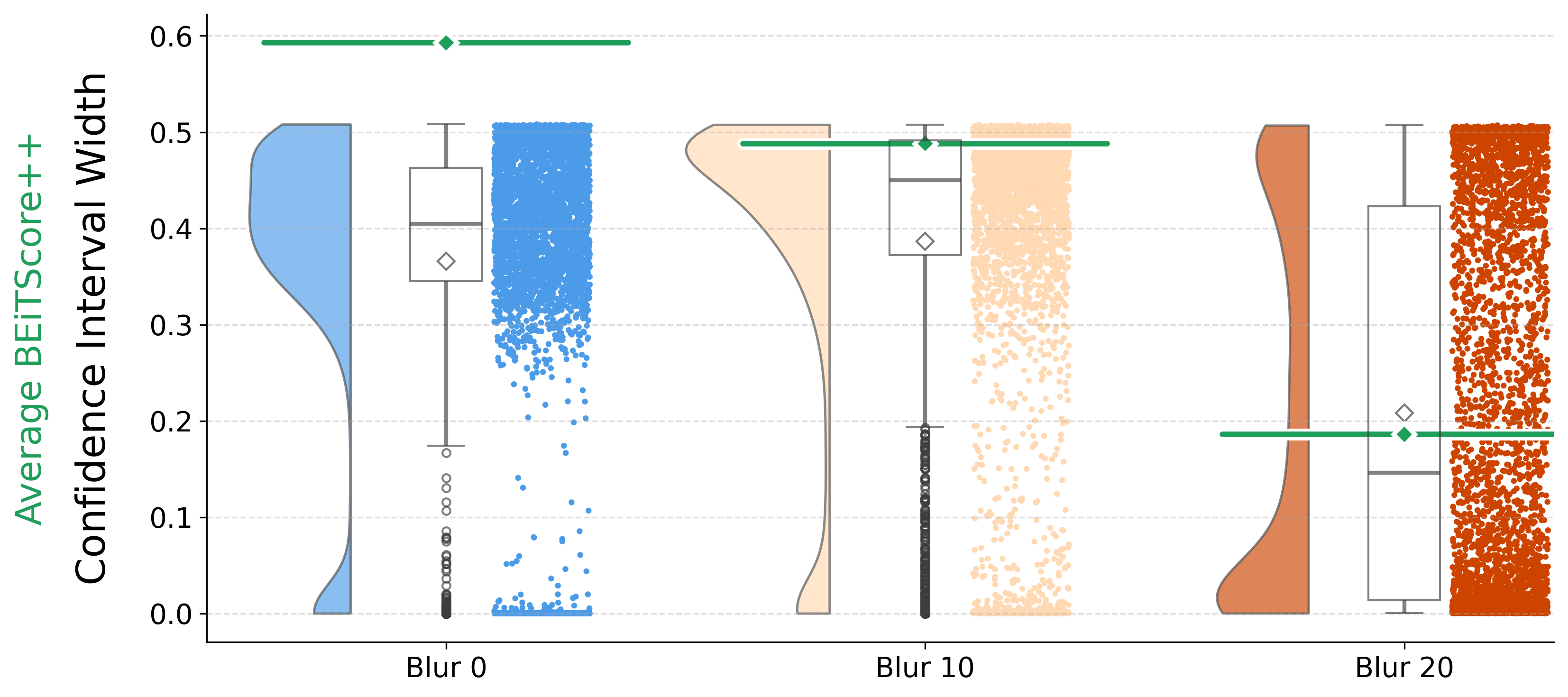}
    \caption{Distribution of 90\% confidence interval widths generated by the base variant of BEiTScore++ as Gaussian blur increases on images from the Polaris test split. The horizontal lines indicate the average median score produced by BEiTScore++ at each blur level.}
    \label{fig:box_plot_input_amb_base}
\end{figure*}

Figure~\ref{fig:box_plot_input_amb_base} shows the confidence interval widths
produced by the base model under increasing blur levels. The results follow the
same pattern observed for the large model: the interval width grows under moderate blur, indicating that the estimator responds appropriately to increasing
input ambiguity. As before, the strongest blur level produces a reduction
in interval width, which we attribute to the model becoming confident that the
image-caption pair is strongly misaligned.

\begin{figure*}[t]
    \centering
    \includegraphics[width=\linewidth]{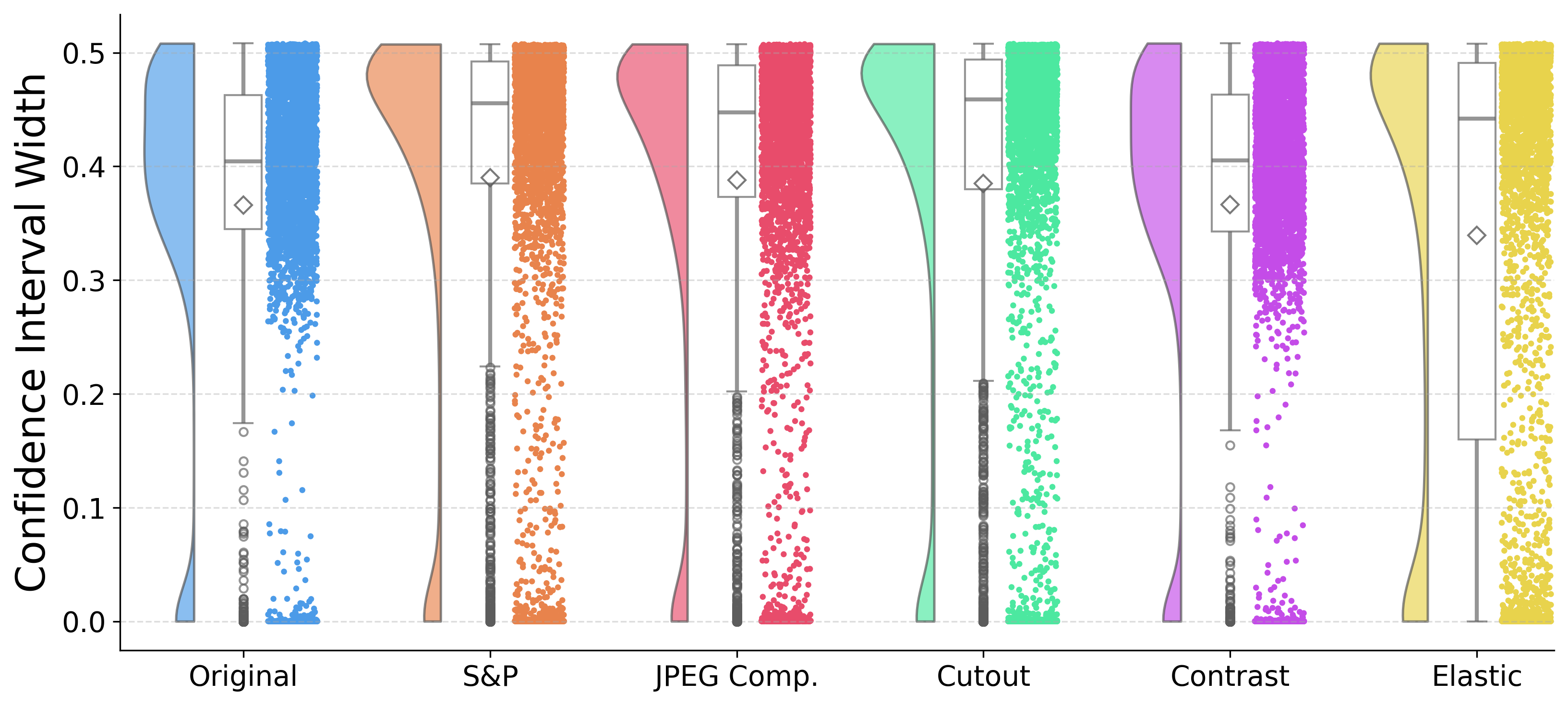}
    \caption{Box plots of confidence interval widths produced by the base-sized BEiTScore++ model, under different types of distortions applied to images from the Polaris test split.}
    \label{fig:box_plot_dist_base}
\end{figure*}

We also evaluate the base model under the broader set of distortions described in Appendix~\ref{app:beyond_blur}, including salt-and-pepper noise, JPEG compression,
random patch cutout, elastic warping, and contrast adjustments. The results are
consistent with the large model, and can be seen in
Figure~\ref{fig:box_plot_dist_base}. Most distortions lead to larger confidence
intervals, while contrast and other color-based
transformations produce little to no change in uncertainty. This suggests that
the base model is likewise robust to purely photometric perturbations and
primarily sensitive to structural corruption of the image.

\subsection{Model Knowledge Gaps}

We next examine whether the base model reflects uncertainty arising from model
knowledge gaps. We compare in-domain datasets containing natural images and realistic scenes, namely IIW, DOCCI, Localized Narratives, and MSCOCO, against out-of-domain artwork datasets, namely Docent and ArtCaps. We again split the evaluation into long-caption and
short-caption settings, to match the caption-length behavior expected by
BEiTScore.

For long captions, we compare IIW, DOCCI, and Localized Narratives against
Docent. For short captions, we compare ArtCaps against a randomly sampled subset
of MSCOCO of equal size.

\begin{figure*}[t]
    \centering
    \includegraphics[width=\linewidth]{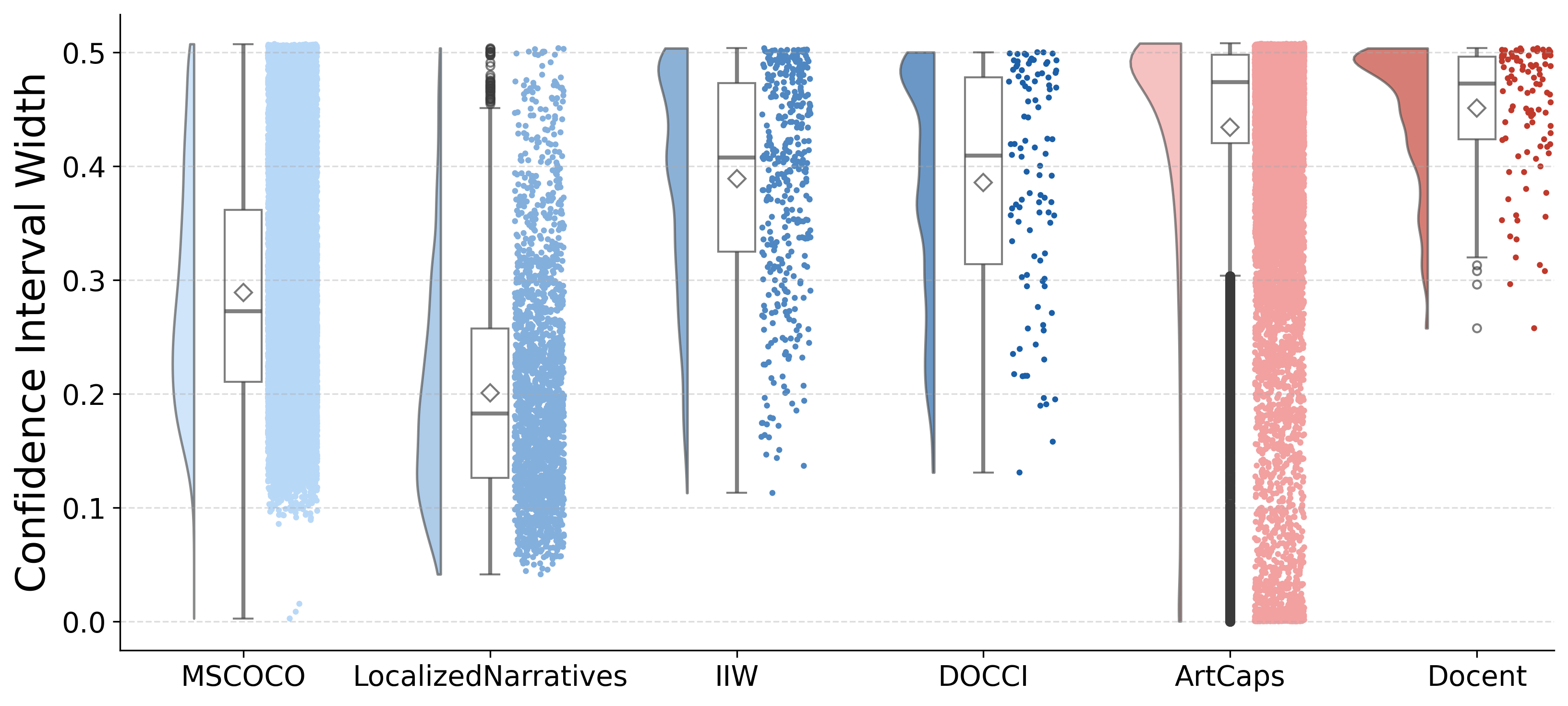}
    \caption{Distribution of 90\% confidence interval widths produced by the base variant of BEiTScore++ for in-domain (natural images) and out-of-domain (artwork) datasets, across both short- and long-caption settings. The color gradient represents caption length, with pastel tones indicating short and direct captions, and with more saturated tones corresponding to datasets with more detailed descriptions.}
    \label{fig:model_gap_base}
\end{figure*}

Figure~\ref{fig:model_gap_base} shows that the base model, like the large model,
assigns substantially higher uncertainty to artwork datasets in both settings.
As for the large model, this increase is consistent with epistemic uncertainty
induced by domain shift, although it may also reflect the greater ambiguity of the artwork data.

\subsection{Marginal Distributions for the BEiTScore++ Predicted Quantiles}
\label{app:quantile-distributions}

\begin{figure*}[htbp]
    \centering
    \includegraphics[width=\textwidth]{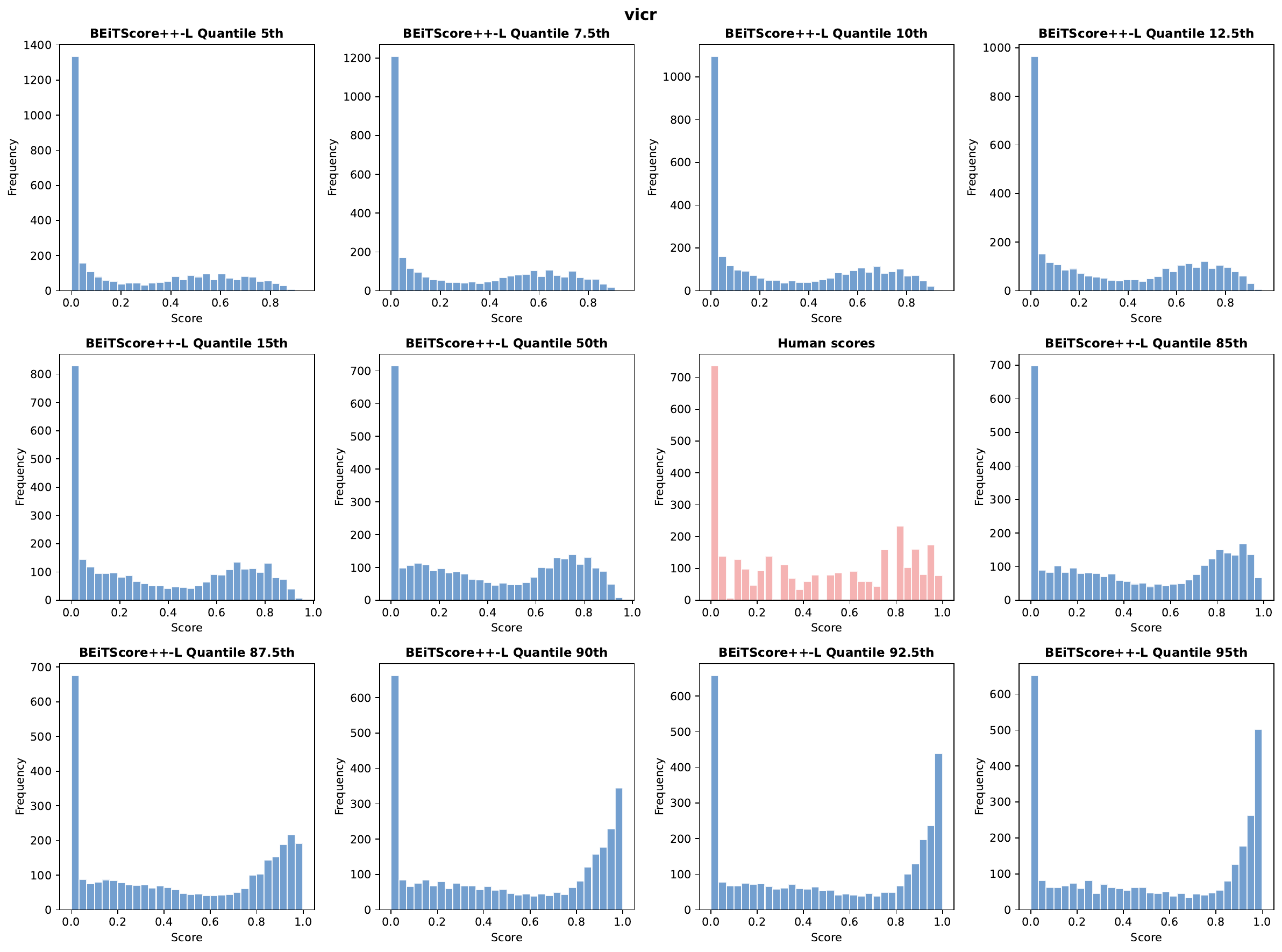}
    \caption{Marginal distributions of the predicted quantiles and the mean
    human ratings for the VICR dataset.}
    \label{fig:quantile-distributions-vicr}
\end{figure*}

\begin{figure*}[htbp]
    \centering
    \includegraphics[width=\textwidth]{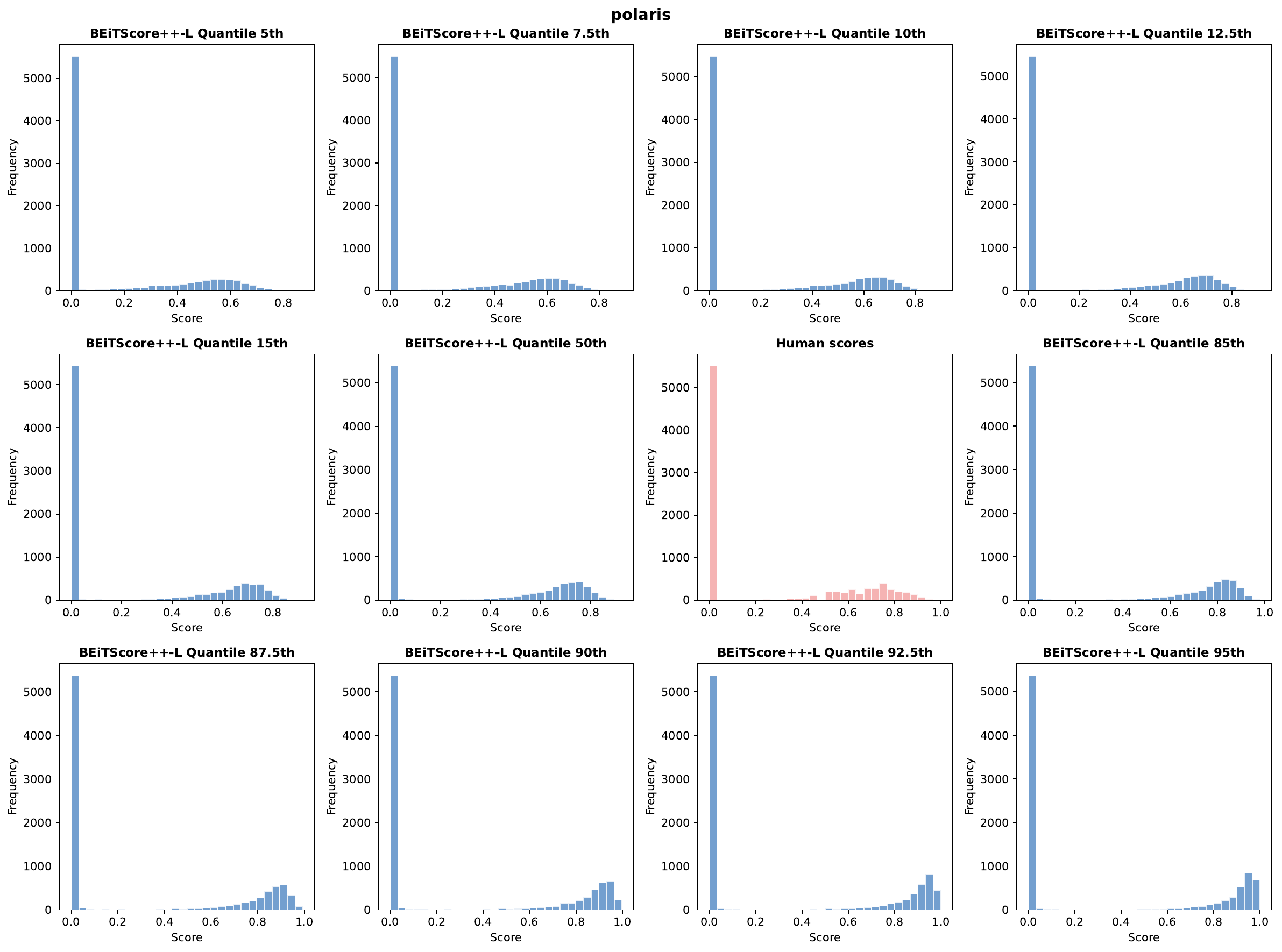}
    \caption{Marginal distributions of the predicted quantiles and the mean
    human ratings for the Polaris dataset.}
    \label{fig:quantile-distributions-polaris}
\end{figure*}

\begin{figure*}[htbp]
    \centering
    \includegraphics[width=\textwidth]{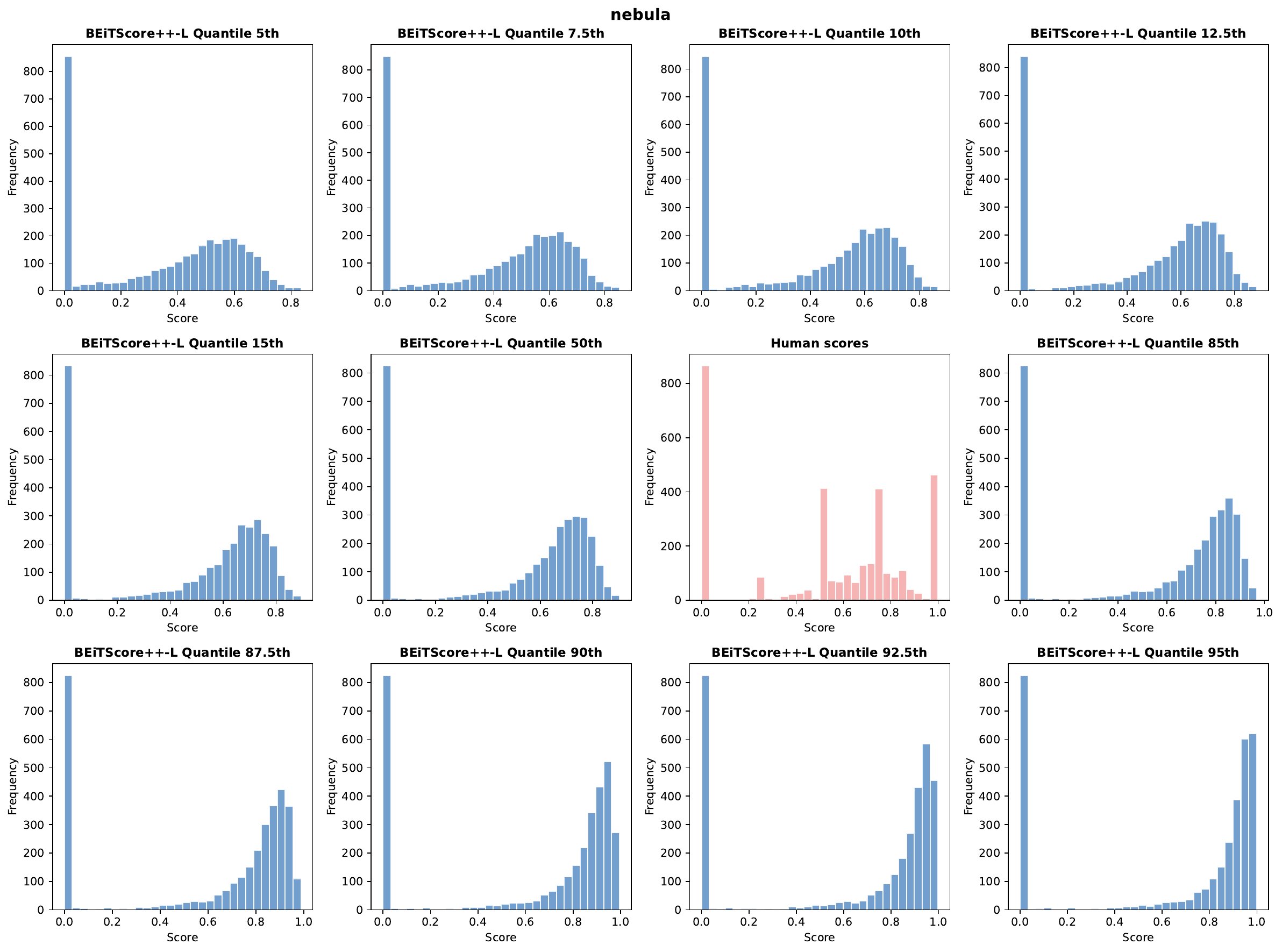}
    \caption{Marginal distributions of the predicted quantiles and the mean
    human ratings for the Nebula dataset.}
    \label{fig:quantile-distributions-nebula}
\end{figure*}

\begin{figure*}[htbp]
    \centering
    \includegraphics[width=\textwidth]{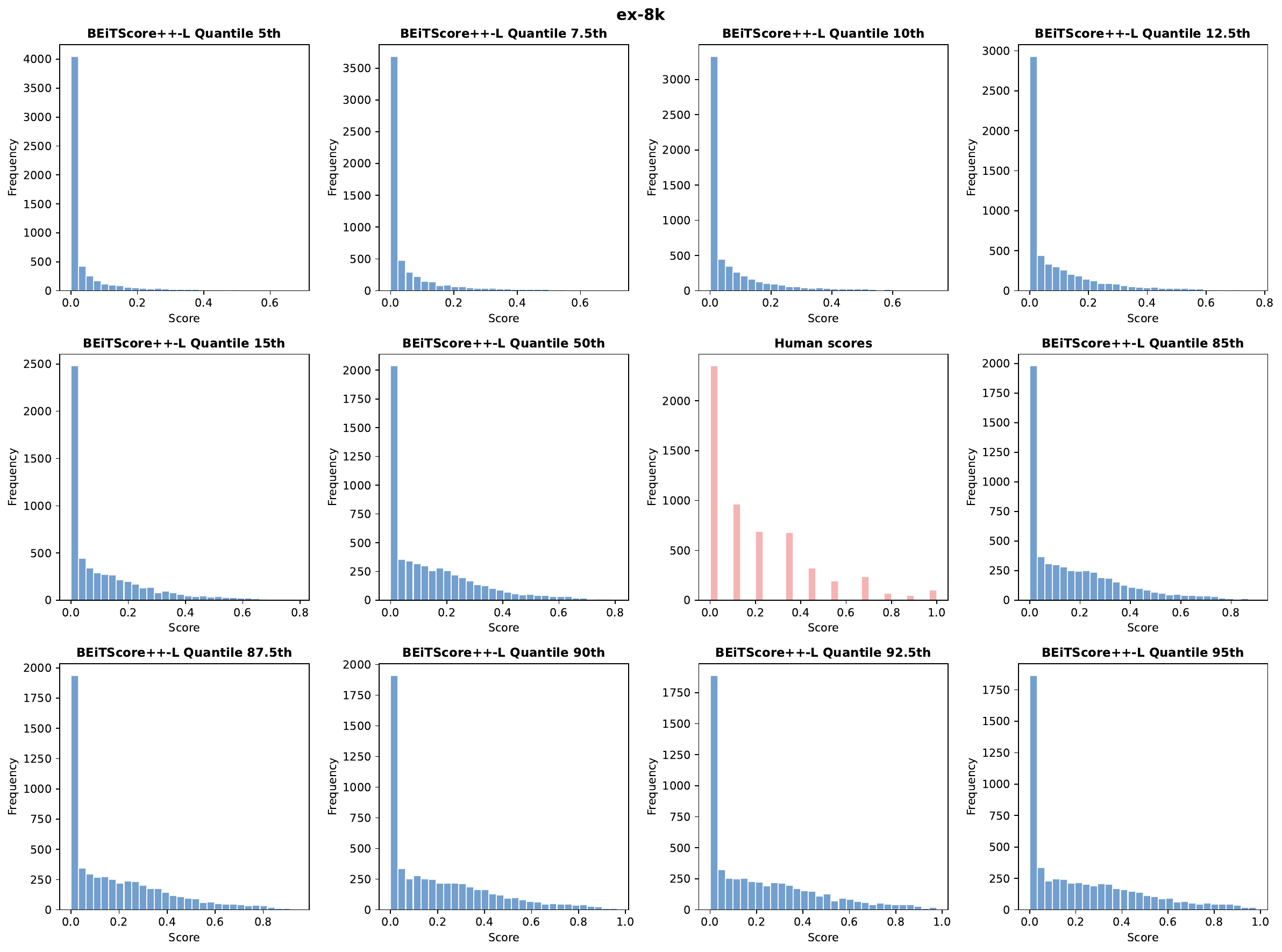}
    \caption{Marginal distributions of the predicted quantiles and the mean
    human ratings for the Ex-8K dataset.}
    \label{fig:quantile-distributions-ex8k}
\end{figure*}

\begin{figure*}[htbp]
    \centering
    \includegraphics[width=\textwidth]{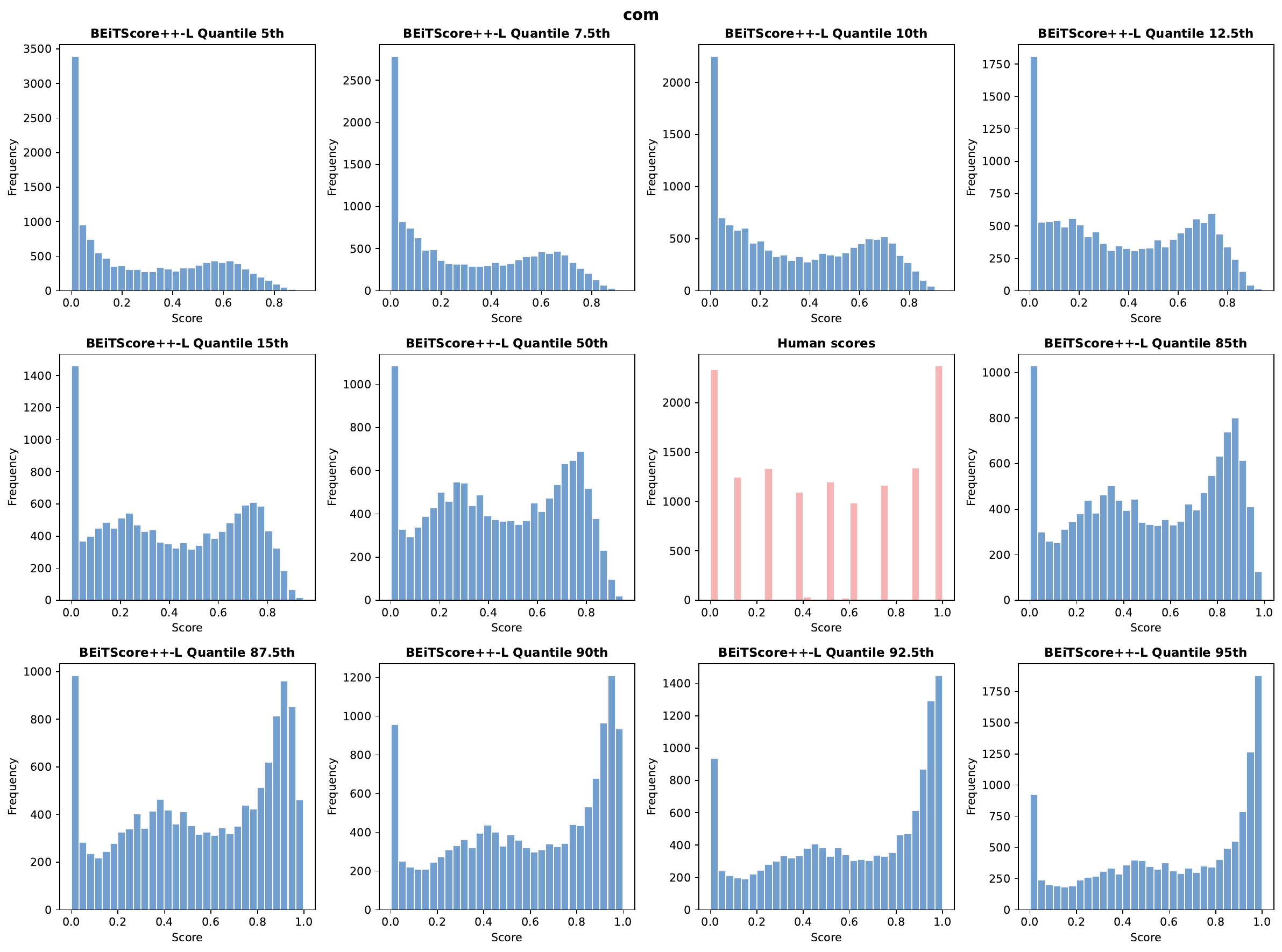}
    \caption{Marginal distributions of the predicted quantiles and the mean
    human ratings for the COM dataset.}
    \label{fig:quantile-distributions-com}
\end{figure*}

This section reports, for every evaluated dataset, the marginal
distribution of each predicted quantile produced by BEiTScore++-L,
together with the distribution of the mean human ratings. Each panel
aggregates the prediction, prior to conformal calibration, over all instances in the corresponding
dataset, thus, the histograms describe how a given quantile level is
distributed across the evaluation set, rather than the interval predicted
for any single caption. The distributions for the VICR, Polaris,
Nebula, Ex-8K, and COM datasets are respectively shown in
Figures~\ref{fig:quantile-distributions-vicr}
to~\ref{fig:quantile-distributions-com}.

The same qualitative behavior is observed across all datasets. The panels
display the expected monotone ordering of the quantile levels. The lower
quantiles, from the 5th to the 12.5th, are heavily concentrated near the
lower end of the score range, with the bulk of their mass in that region.
As the quantile level increases, the mass progressively shifts to the right
and spreads over the full support of the score. The 50th quantile, which
corresponds to the median prediction, covers the entire range and its shape
closely tracks the distribution of the mean human ratings shown in the same
figure. The upper quantiles, from the 85th to the 95th, are shifted towards
the higher end of the range and concentrate near the upper bound, mirroring
the behavior of the lower quantiles at the opposite extreme.

Since the outer quantiles bound the widest intervals in the nested family, they tend to accumulate near the extremes of the score range, while the median remains aligned with the distribution of the human ratings, indicating that the central prediction follows the scale of the human judgments. Quantile crossing is excluded by the monotone construction, so the ordering of the quantile levels holds for every instance, and not only for the marginal distributions shown in these figures. These observations hold across the five datasets, which indicates that the behavior is a property of the model rather than an artifact of any particular evaluation set.

\subsection{Qualitative Examples}
\label{app:qualitative-examples}

This section presents qualitative examples of the predictions produced by
BEiTScore++-L for every evaluated dataset. For each example, the corresponding figure
shows the input image, the candidate caption, the median score predicted by
the model, and the 90 percent prediction interval obtained after applying
the calibrated scaling and bias correction. The mean human rating is
reported alongside each prediction so that the point estimate and the
predicted interval can be compared directly against the ground truth. The
examples were sampled at random from each evaluation set, without any
selection based on the quality of the prediction, so that the figures
reflect the typical behavior of the model rather than a curated subset.

The examples illustrate the general behavior discussed in the main text.
In the majority of cases the mean human rating falls within the predicted
90 percent interval, which is consistent with the calibrated coverage
reported for the model. The width of the interval also varies with the
difficulty of the instance, with wider intervals appearing for captions
whose agreement with the image is more ambiguous and narrower intervals for
clear cases at the extremes of the score range. Cases in which the human
rating falls outside the interval are also present, as expected for a
method that targets a nominal coverage level rather than exact containment. These cases are distributed across the datasets without any obvious
systematic pattern.

\begin{figure*}[p]
    \centering
    \includegraphics[width=\textwidth]{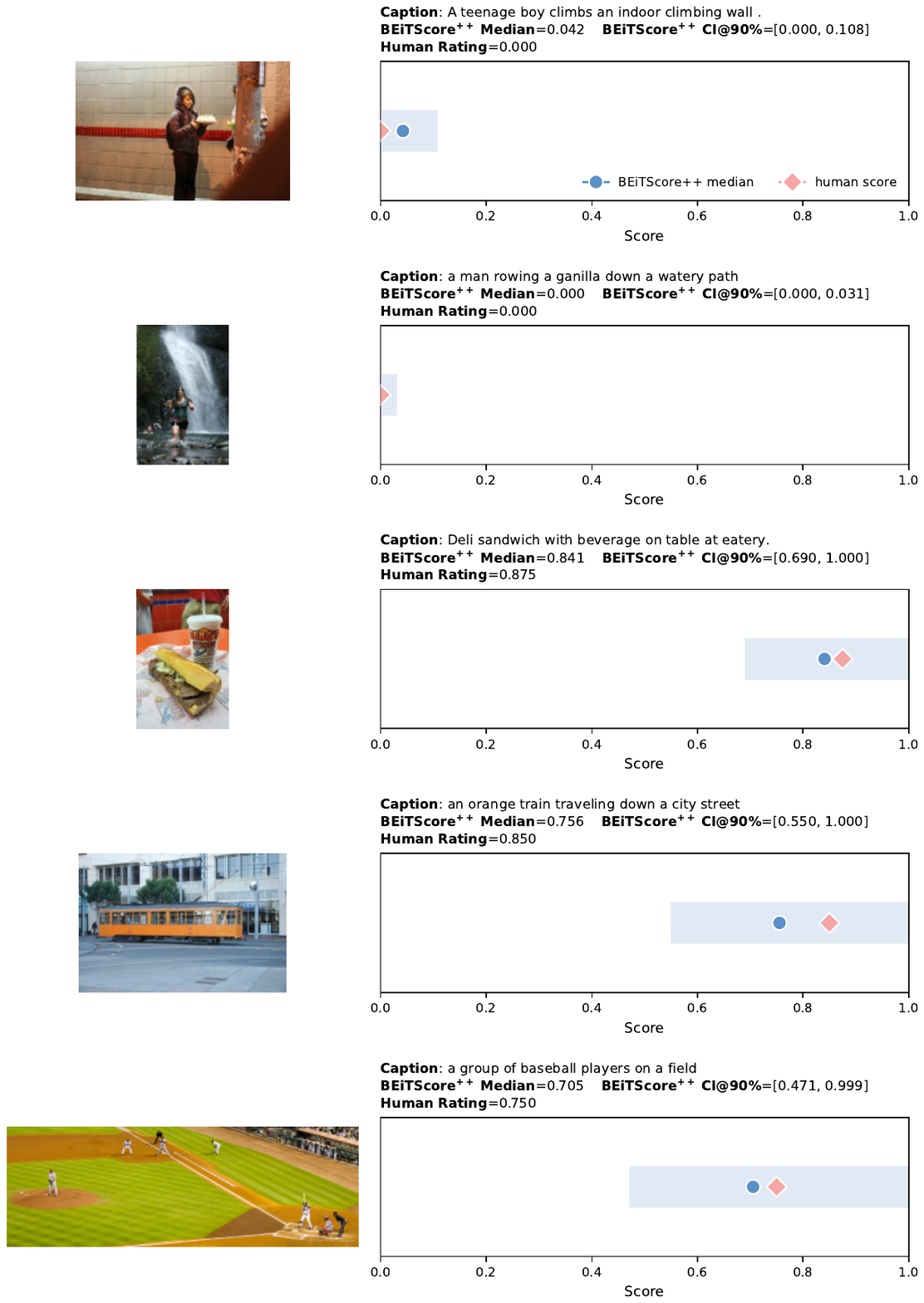}
    \caption{Qualitative examples for the predictions of BEiTScore++-L
    on the VICR dataset. Each row shows the input image, the candidate
    caption, the predicted median and 90 percent interval, and the mean
    human rating.}
    \label{fig:qualitative-vicr-p1}
\end{figure*}

\begin{figure*}[p]
    \centering
    \includegraphics[width=\textwidth]{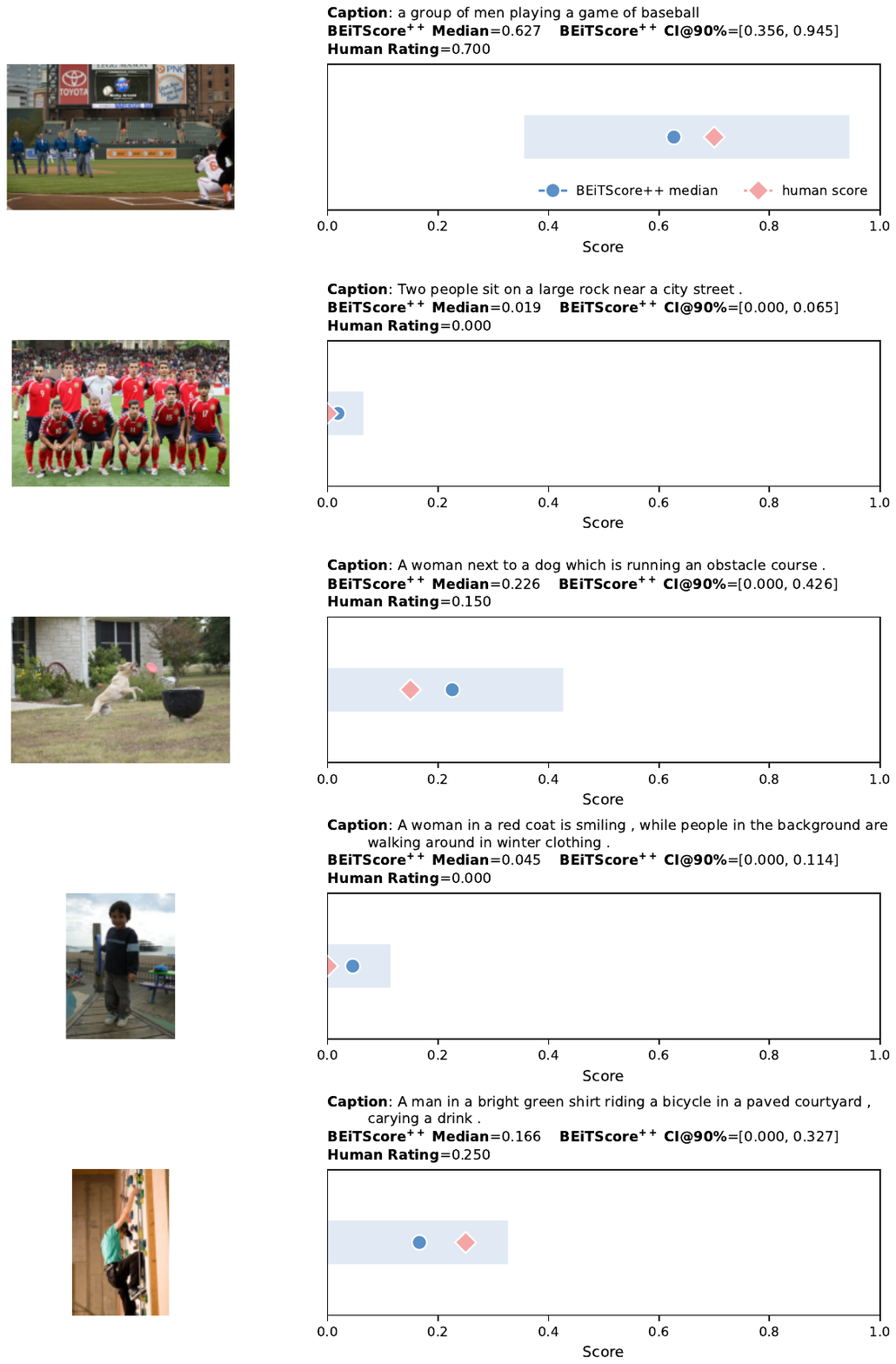}
    \caption{Additional qualitative examples for the predictions of
    BEiTScore++-L on the VICR dataset.}
    \label{fig:qualitative-vicr-p2}
\end{figure*}

\begin{figure*}[p]
    \centering
    \includegraphics[width=\textwidth]{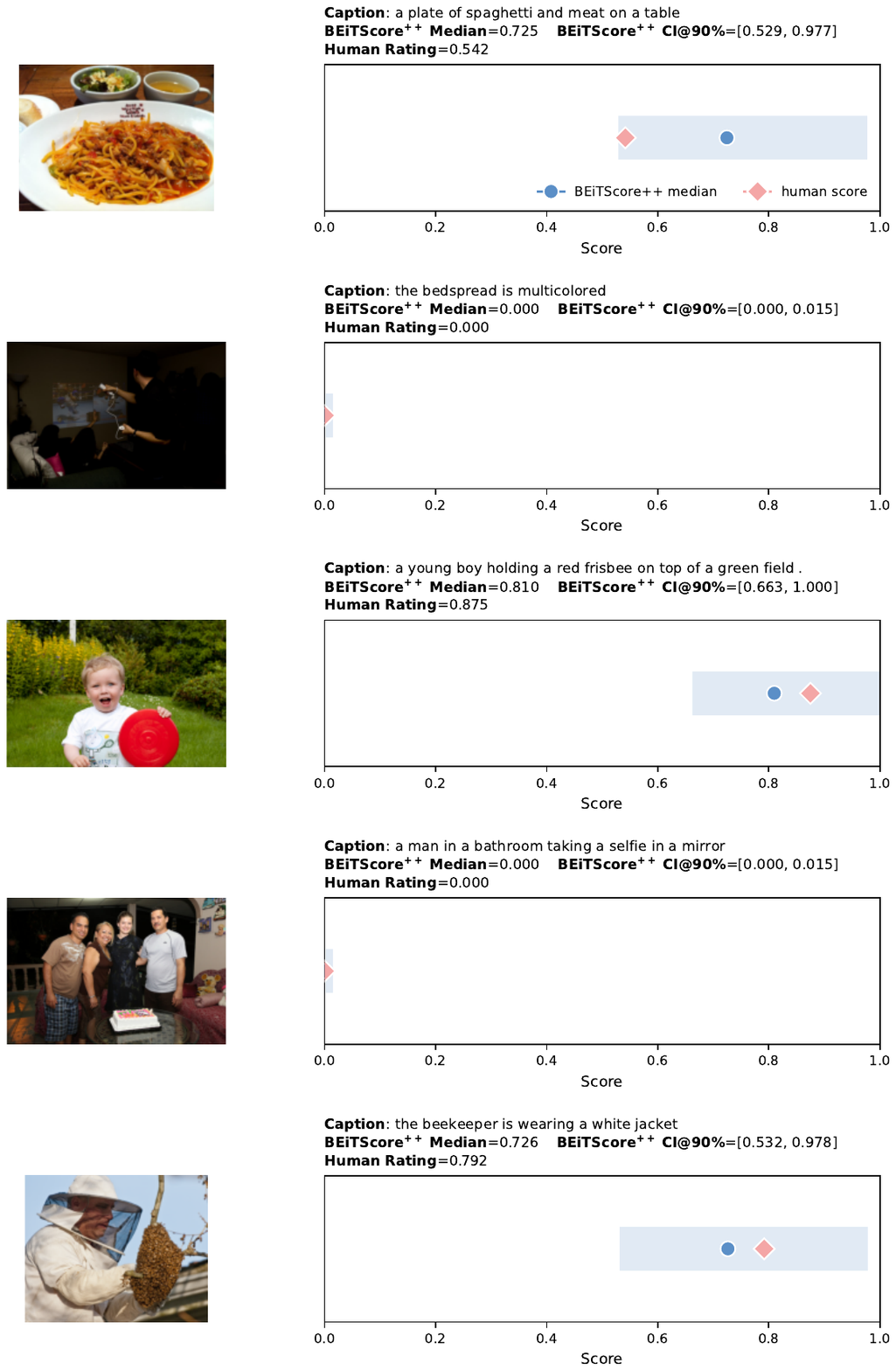}
    \caption{Qualitative examples for the predictions of BEiTScore++-L
    on the Polaris dataset. Each row shows the input image, the candidate
    caption, the predicted median and 90 percent interval, and the mean
    human rating.}
    \label{fig:qualitative-polaris-p1}
\end{figure*}

\begin{figure*}[p]
    \centering
    \includegraphics[width=\textwidth]{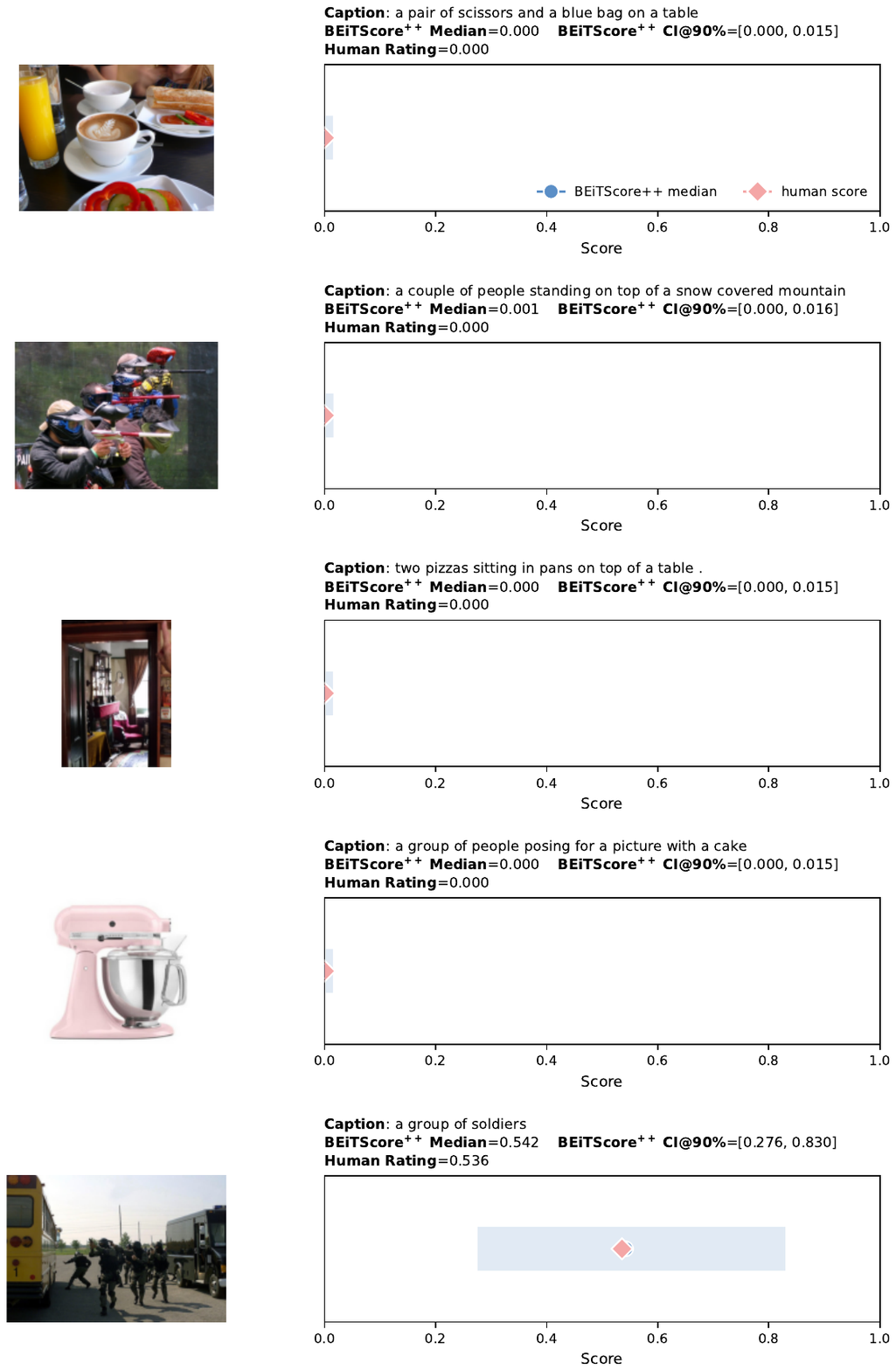}
    \caption{Additional qualitative examples for the predictions of
    BEiTScore++-L on the Polaris dataset.}
    \label{fig:qualitative-polaris-p2}
\end{figure*}

\begin{figure*}[p]
    \centering
    \includegraphics[width=\textwidth]{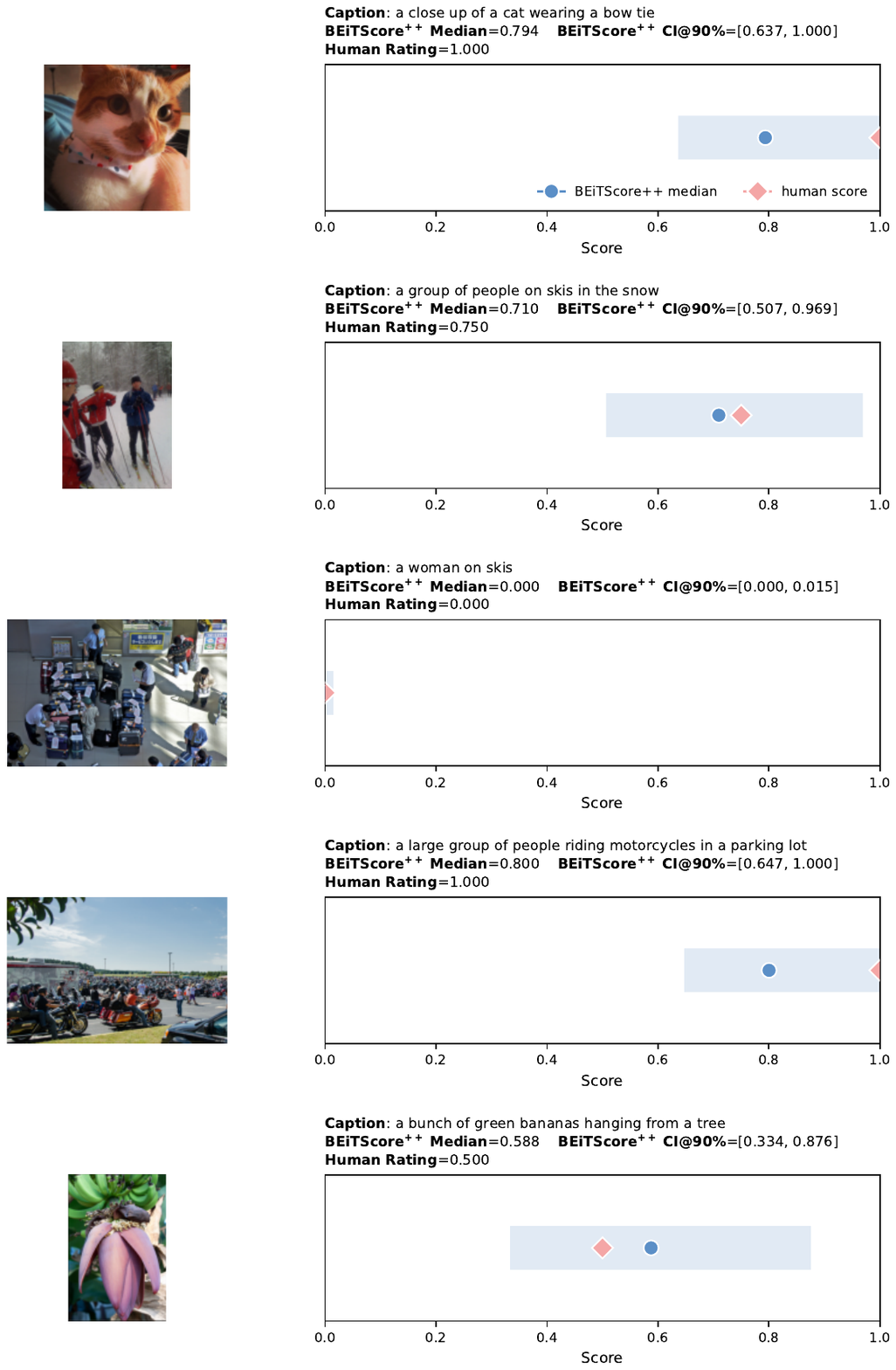}
    \caption{Qualitative examples for the predictions of BEiTScore++-L
    on the Nebula dataset. Each row shows the input image, the candidate
    caption, the predicted median and 90 percent interval, and the mean
    human rating.}
    \label{fig:qualitative-nebula-p1}
\end{figure*}

\begin{figure*}[p]
    \centering
    \includegraphics[width=\textwidth]{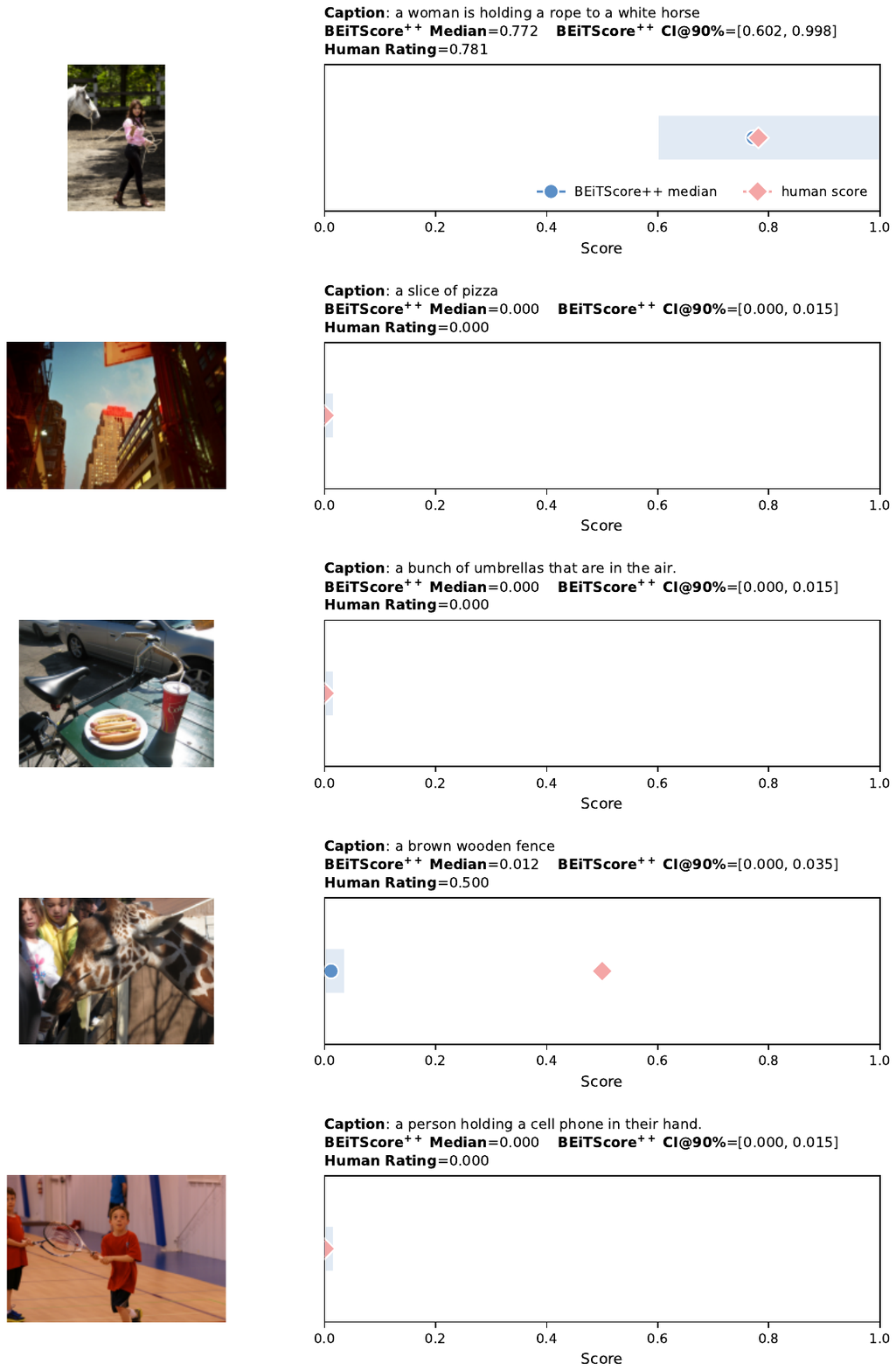}
    \caption{Additional qualitative examples for the predictions of
    BEiTScore++-L on the Nebula dataset.}
    \label{fig:qualitative-nebula-p2}
\end{figure*}

\begin{figure*}[p]
    \centering
    \includegraphics[width=\textwidth]{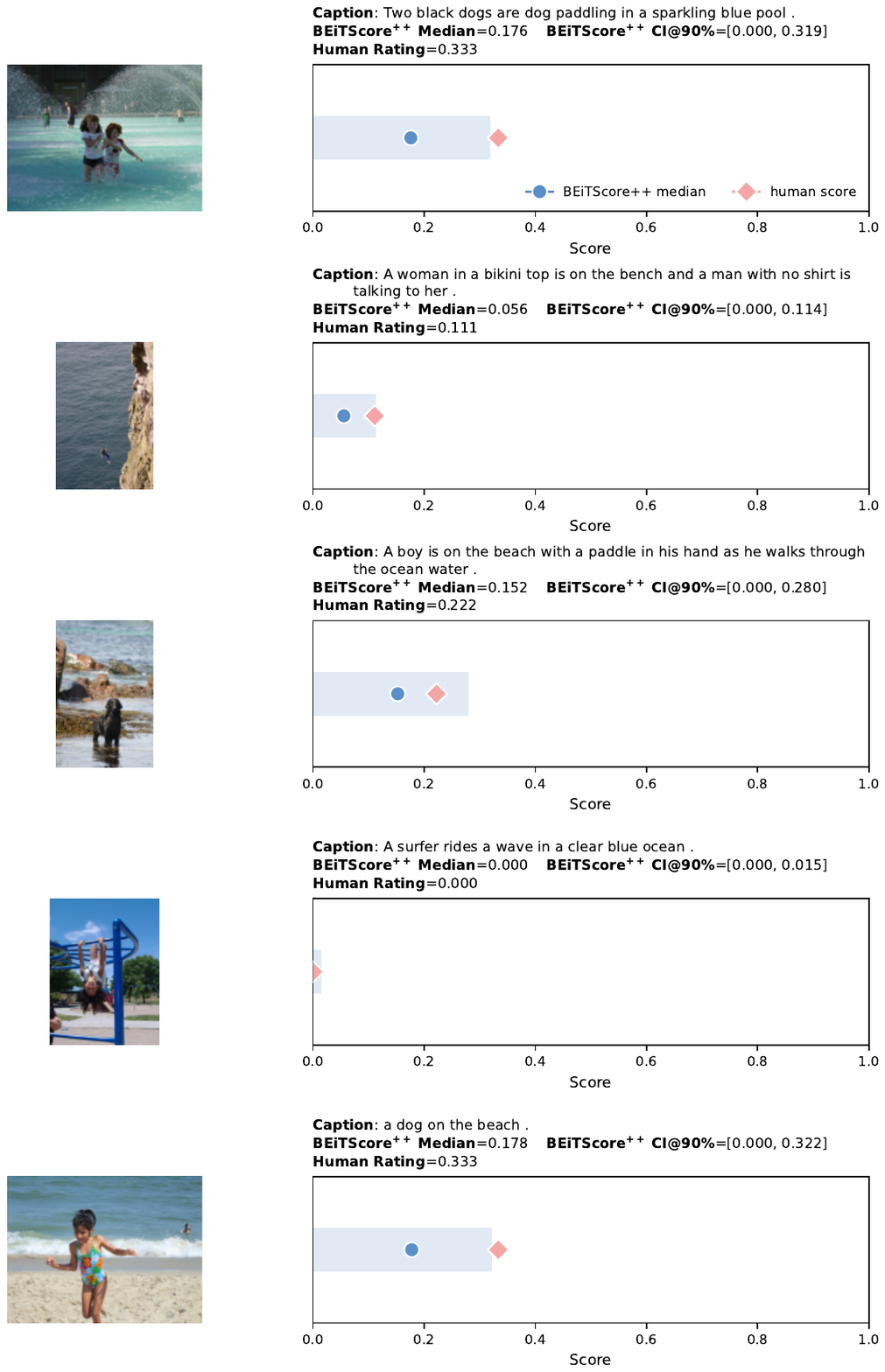}
    \caption{Qualitative examples for the predictions of BEiTScore++-L
    on the Ex-8K dataset. Each row shows the input image, the candidate
    caption, the predicted median and 90 percent interval, and the mean
    human rating.}
    \label{fig:qualitative-ex8k-p1}
\end{figure*}

\begin{figure*}[p]
    \centering
    \includegraphics[width=\textwidth]{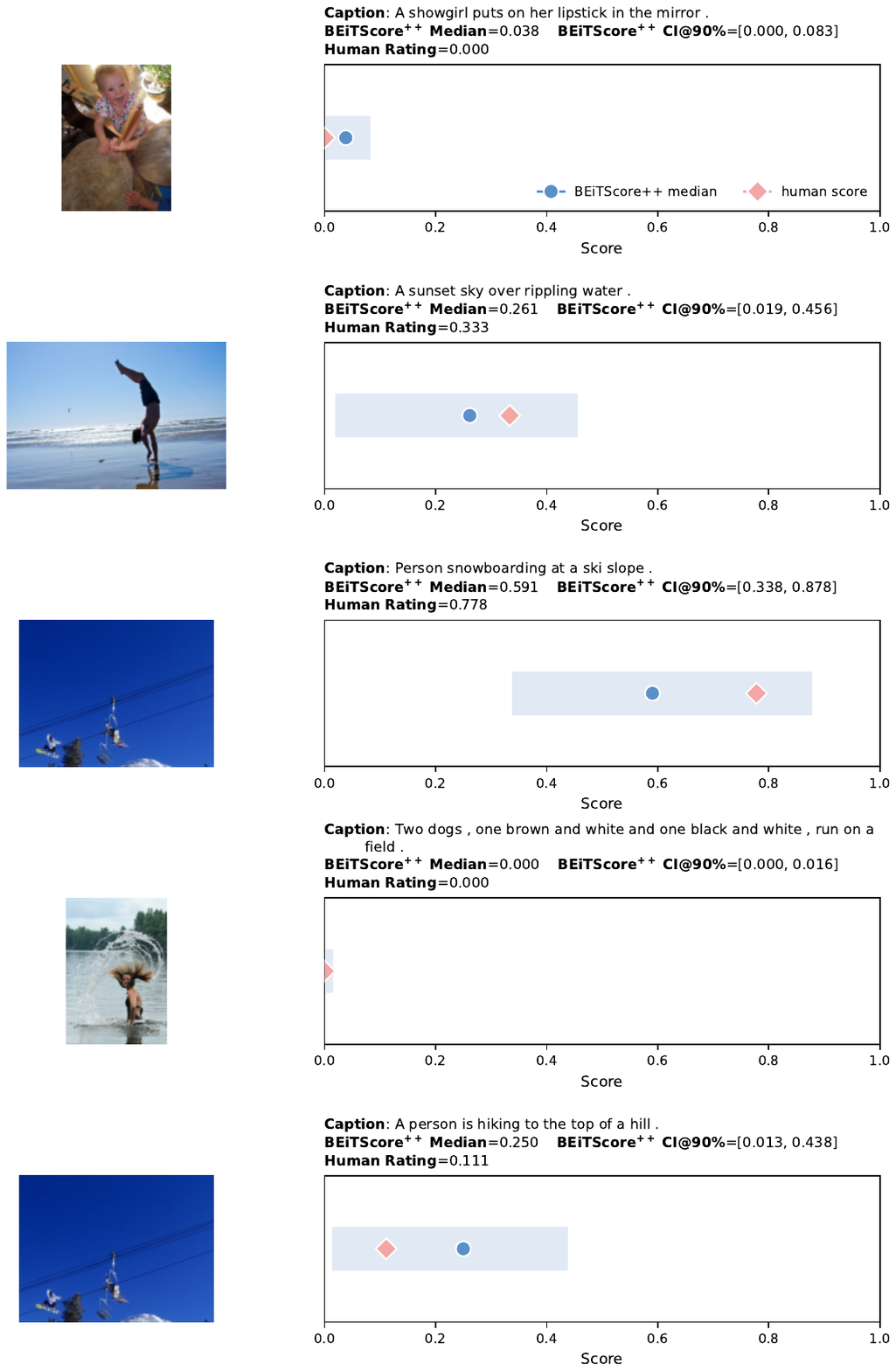}
    \caption{Additional qualitative examples for the predictions of
    BEiTScore++-L on the Ex-8K dataset.}
    \label{fig:qualitative-ex8k-p2}
\end{figure*}

\begin{figure*}[p]
    \centering
    \includegraphics[width=\textwidth]{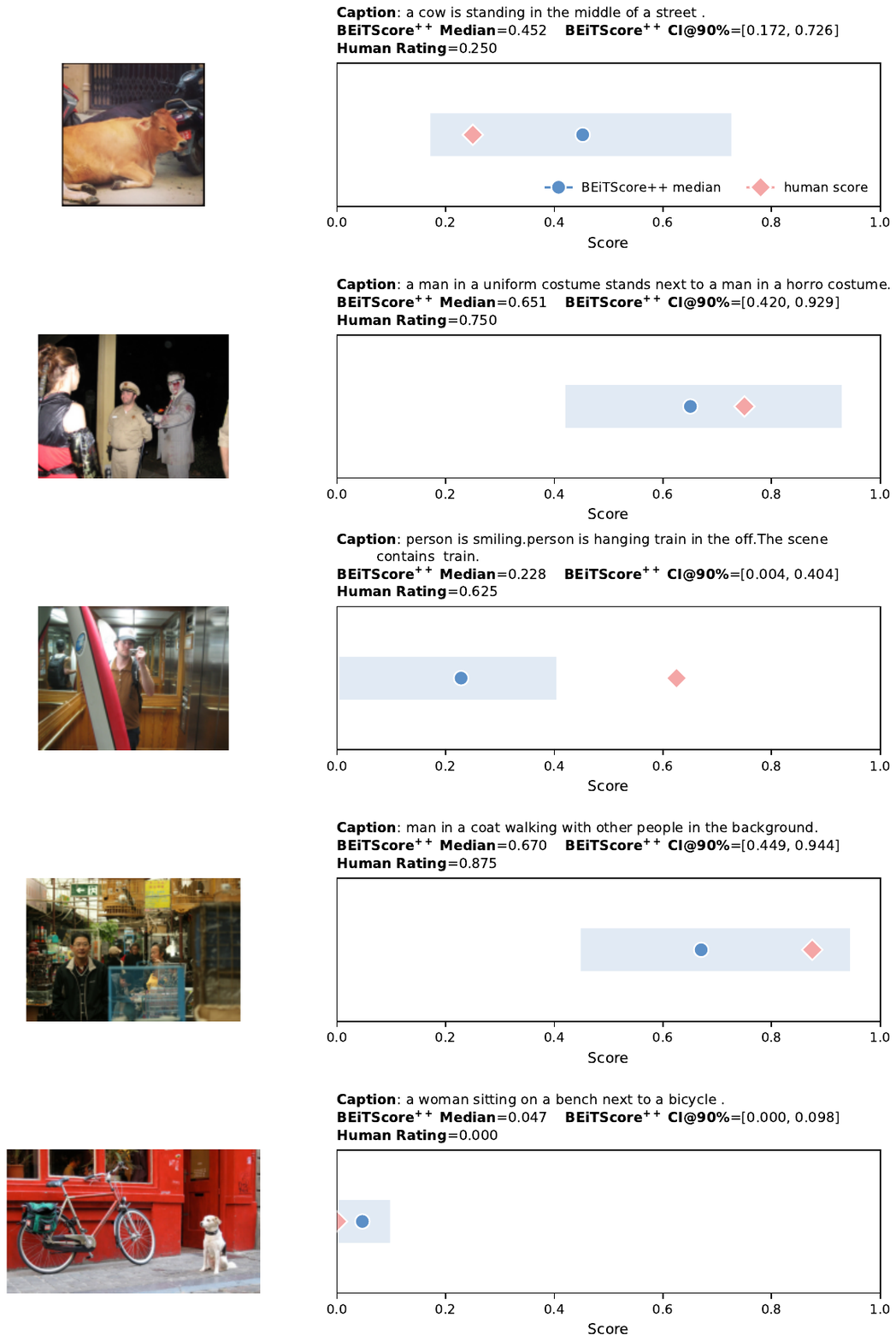}
    \caption{Qualitative examples for the predictions of BEiTScore++-L
    on the COM dataset. Each row shows the input image, the candidate
    caption, the predicted median and 90 percent interval, and the mean
    human rating.}
    \label{fig:qualitative-com-p1}
\end{figure*}

\begin{figure*}[p]
    \centering
    \includegraphics[width=\textwidth]{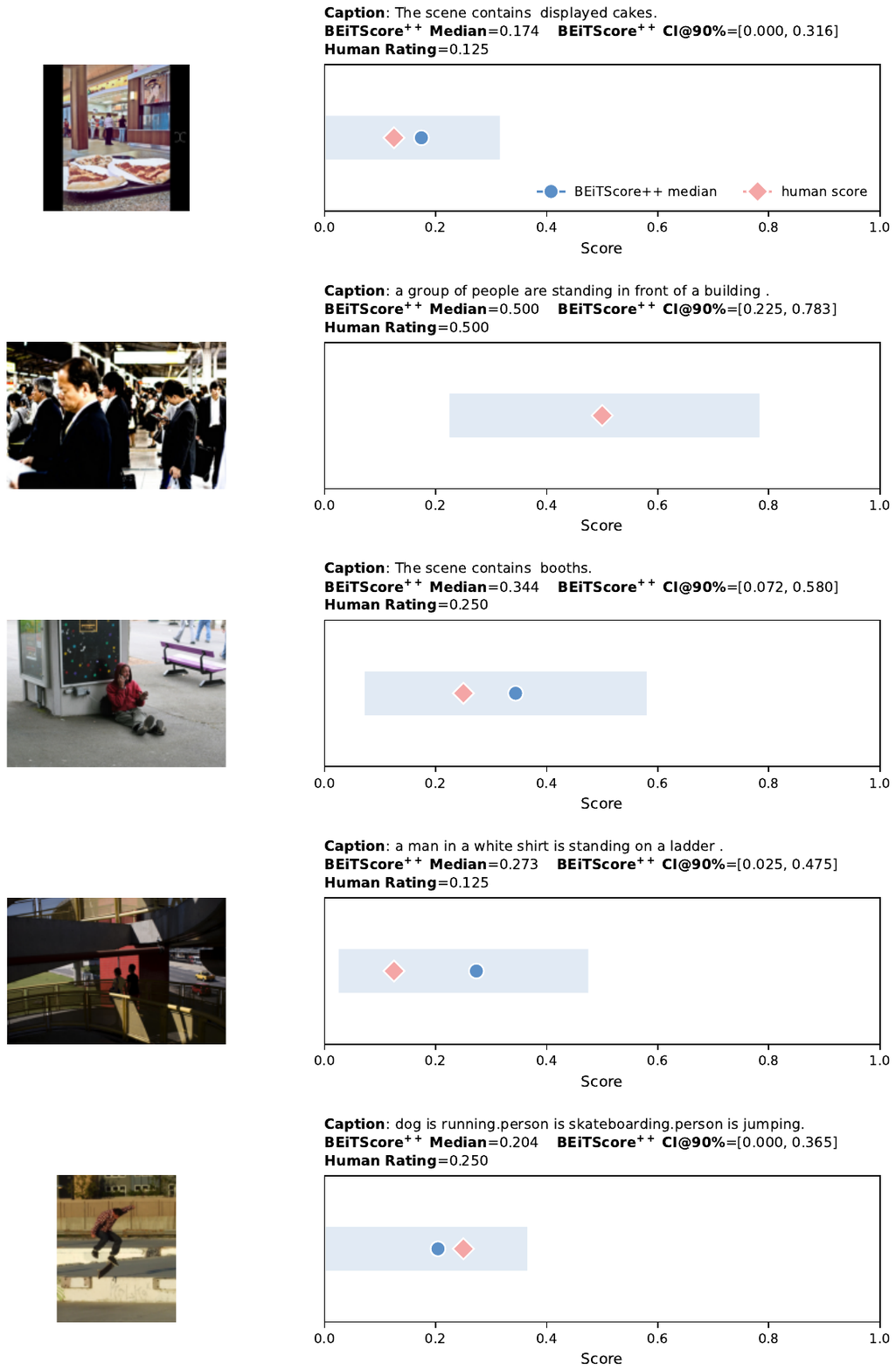}
    \caption{Additional qualitative examples for the predictions of
    BEiTScore++-L on the COM dataset.}
    \label{fig:qualitative-com-p2}
\end{figure*}

Multiple examples for each of the VICR, Polaris, Nebula, Ex-8K, and COM datasets, are
shown in Figures~\ref{fig:qualitative-vicr-p1}
to~\ref{fig:qualitative-com-p2}.

\end{document}